\RequirePackage{fix-cm}
\documentclass[twocolumn]{svjour3_arxiv}          %
\smartqed  %
\usepackage{graphicx}
\usepackage{eso-pic}
\usepackage{xspace}
\usepackage{url}
\usepackage{times}
\usepackage{epsfig}
\usepackage{graphicx}
\usepackage{amsmath}
\usepackage{amssymb}
\usepackage{bm}
\usepackage{booktabs}
\usepackage{comment}
\usepackage{color}
\usepackage{enumitem}
\usepackage{multirow}
\usepackage{xcolor,colortbl}
\usepackage{bbm}
\usepackage[numbers]{natbib}
\usepackage{float}

\definecolor{Gray}{gray}{0.85}
\definecolor{LightCyan}{rgb}{0.88,1,1}
\newcolumntype{C}{>{\columncolor{Gray}}c}

\usepackage[pagebackref=true,breaklinks=true,letterpaper=true,colorlinks,bookmarks=false]{hyperref}

\begin{document}

\title{Modeling Spatio-Temporal Human Track Structure for Action Localization
}

\author{Guilhem Ch\'{e}ron         \and
        Anton Osokin \and
        Ivan Laptev \and
        Cordelia Schmid
}

\institute{
	Guilhem Ch\'{e}ron~\textsuperscript{*}~\textsuperscript{$\dagger$} \at
    \email{guilhem.cheron@inria.fr} 
    \and
    Anton Osokin~\textsuperscript{$\ddagger$}~\textsuperscript{*} \at
    \email{aosokin@hse.ru} 
    \and
    Ivan Laptev~\textsuperscript{*} \at
    \email{ivan.laptev@inria.fr} 
    \and
    Cordelia Schmid~\textsuperscript{$\dagger$} \at
    \email{cordelia.schmid@inria.fr
    \and
    \textsuperscript{*}~Inria Paris, ENS, CNRS, UMR 8548 (Willow team)\\
	\textsuperscript{$\dagger$}~Inria Grenoble, LJK, CNRS, Univ. Grenoble Alpes (Thoth team) \\
	\textsuperscript{$\ddagger$}~National Research University Higher School of Economics, Moscow} 
}

\maketitle
\begin{abstract}
This paper addresses spatio-temporal localization of human actions in video. In order to 
localize actions in time, we propose a recurrent localization network (RecLNet)
designed to model the temporal structure of actions on the level of
person tracks.  Our model is trained to simultaneously recognize
and localize action classes in time and is
based on two layer gated recurrent units (GRU) applied separately to two streams, i.e.~appearance and optical flow streams.
When 
used together with state-of-the-art person detection and tracking, our
model is shown to improve substantially spatio-temporal action localization in
videos. The gain is shown to be mainly due to improved temporal localization.
We evaluate our method on two recent datasets for spatio-temporal
action localization, UCF101-24 and DALY, demonstrating a  significant
improvement of the state of the art.  
\end{abstract}

\vspace{-5mm}\section{Introduction}

\global\csname @topnum\endcsname 0
\global\csname @botnum\endcsname 0
\begin{figure}
    \includegraphics[trim =0mm 50mm 0mm 0mm,clip, scale=0.168]{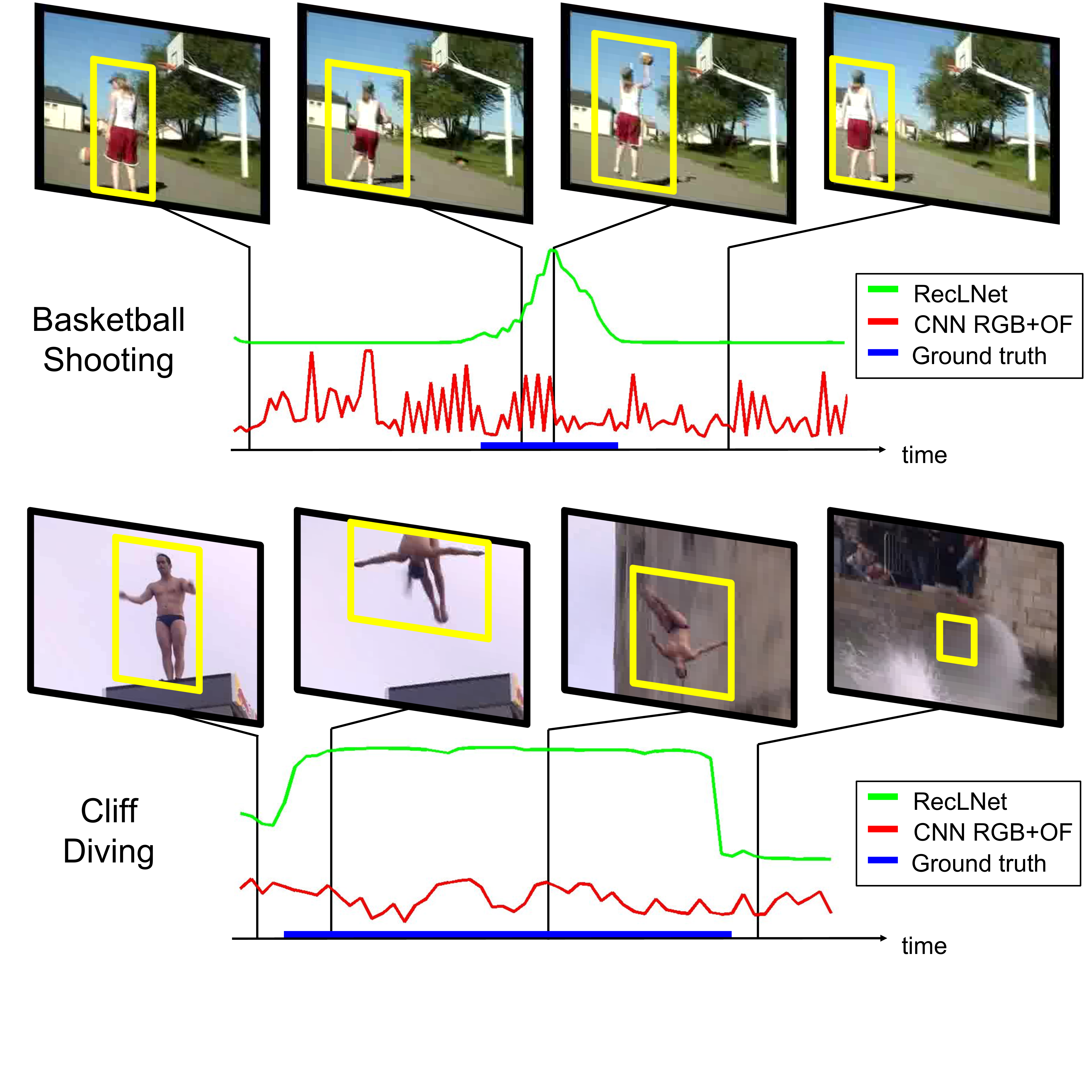} %
    \caption{Spatio-temporal action localization using a CNN baseline
      (red) and our RecLNet (green) both applied on the level
      of person tracks. Our approach provides accurate temporal boundaries when the action happens.\vspace{-.2cm}} 
    \label{fig:teaserlstm}
\end{figure}

\vspace{-2mm}Successful action recognition will help us drive our cars, 
prevent crime, search our video collections and will eventually enable robots to serve us at home.
Such applications require action localization, i.e.~identifying when the action happens and who is performing the action.
Most of the current methods and benchmarks for action recognition,
however, only address action
classification~\cite{souza2016,simonyan2014two}, i.e.~assuming
temporally segmented action intervals as input.

Identifying the beginning and the end of an action naturally suggests
the need of temporal models for video sequences. Sequence models have
previously been explored for sound, speech and text understanding. In
particular, recurrent neural network models (RNNs) have recently shown
success for speech recognition~\cite{dahl2012context} and text
generation~\cite{sutskever2011generating} as well as for image and
video
captioning~\cite{donahue2015,karpathy2015deep,vinyals2015show}. RNNs
have also been explored for action classification in
video~\cite{donahue2015,Ng_2015_CVPR}, but have shown limited
improvements for this task so far.

Action classification may not require sophisticated temporal models if
classes can be distinguished solely by the presence of action-specific
features. On the other hand, if the structure of the video is required
as the output, explicit spatio-temporal models of the video can be
beneficial. Recent
work~\cite{liu2016,ma2016,singh2016actionRNN,yeung2016,yuan2016} has
indeed shown improvements in temporal action localization achieved
with recurrent models of video sequences. Here we develop and investigate recurrent models for
spatio-temporal action localization.

Our goal is to localize the acting person in the video frame and to identify temporal boundaries of corresponding actions.
To this end, we propose a recurrent localization network (RecLNet) with gated recurrent units (GRU)~\cite{cho2014learning} for modeling actions on the level of person tracks. 
Our method starts by person detection and tracking similar to~\cite{actiontubes,peng2016multi,saha2016deep,Weinzaepfel_2015_ICCV}. %
Differently to previous work, we train our RecLNet to score actions and detect temporal boundaries within each person track (see Figure~\ref{fig:teaserlstm}). This scoring is achieved by two-stream recurrent units exploiting appearance and motion Fast-RCNN~\cite{Girshick_2015_ICCV} features pooled from person boxes, while final detections are obtained using our simple and effective temporal localization method composed of filtering and thresholding.
We provide a thorough experimental evaluation and analyze the impact of recurrence on spatio-temporal action localization by making the following contributions: 
\begin{itemize}
\item we show spatio-temporal action localization improvement supported by our RecLNet trained on a track-level and compare different standard and recurrent architectures;
\item we empirically diagnose \emph{temporal} localization as being a weakness of existing methods which our method is able to correct;
\item our method is complementary to most recent works~\cite{kalogeiton17iccv} mainly focusing on increasing spatial boxes precision and we identify the spatial aspect as being our principal room for improvement;
\item results are reported on the two largest datasets for our task,
namely UCF101-24 detection~\cite{soomro2012ucf101} and DALY~\cite{weinzaepfel2016towards},
for both of these datasets our method results in significant
improvements over the state of the art.
\end{itemize}
The rest of the paper is organized as follows. Section~\ref{sec:RL} reviews the related work on action classification, temporal and spatio-temporal localization. Section~\ref{sec:method} introduces our RecLNet model, its architecture and our threshold temporal localization technique. Section~\ref{sec:experiments} presents our experimental and qualitative results. Section~\ref{sec:conclusion} finally draws conclusions.
\vspace{-0.6cm}

\section{Related work}
\label{sec:RL}

Our work is mostly related to methods for human action classification, temporal action localization and spatio-temporal action detection in video.

\noindent{\bf Action classification.} The majority of action recognition methods targets clip-level video classification, i.e.~the assignment of video clips to a closed set of action classes.
Recent datasets for this task include UCF-101~\cite{soomro2012ucf101}, HMDB~\cite{HMDB51}, ActivityNet~\cite{heilbron2015activitynet} and Sports-1M~\cite{karpathy2014sports1m}.
Local space-time features such as HOG, HOF and IDT~\cite{Laptev08,wang2013action} have shown initial progress for this task.
More recently CNN and RNN-based approaches have been investigated to learn video representations for action recognition.
A combination of motion and appearance information learned by two separate CNN networks has been proposed in~\cite{simonyan2014two}.
Alternative methods based on spatio-temporal convolutions have been studied in~\cite{Ji10,taylor_convgrbm} and more recently using the C3D~\cite{tran2015c3d} and I3D~\cite{CarreiraZ17} architectures. Such method~\cite{CarreiraZ17} takes advantage of transfer learning by training 3D architectures on a large video datasets~\cite{KayCSZHVVGBNSZ17}.
Several works used RNNs for aggregating video information along time~\cite{baccouche2011sequential,donahue2015,Ng_2015_CVPR,pigou2017gesture,singh2016actionRNN,srivastava2016lstm}.
Currently best performing action classification methods combine IDT features with CNN-based representations of motion and appearance ~\cite{souza2016,feichtenhofer2016nips,feichtenhofer2016cvpr,varol16} or pure 3D CNN architecture~\cite{CarreiraZ17}.
RNNs have shown promise for the task of gesture recognition in~\cite{pigou2017gesture} but did not show significant improvements for the more general task of action classification.
In this work, we compute appearance and motion CNN features extracted on the level of person tracks and use them as input to our RecLNet for action localization.

\noindent{\bf Temporal localization.} Temporal action localization aims both to classify and identify temporal extents of actions in longer video clips. Recent datasets for this task include THUMOS~\cite{idrees2016thumos}, 
Activity-Net~\cite{heilbron2015activitynet} and MPII Cooking~\cite{rohrbach15ijcv}.
Methods for joint action classification and temporal segmentation have explored dynamic programming~\cite{hoai2011joint} and temporal grammars~\cite{pirsiavash2014parsing}.
More recently several RNN-based methods have shown gains for action localization in~\cite{bagautdinov2017,ma2016,singh2016actionRNN,yeung2016,yuan2016}.
For example, \cite{ma2016,singh2016actionRNN,yuan2016} explore variants of LSTM to improve per-frame action classification whereas the method in~\cite{yeung2016} learns to directly predict action boundaries.
An alternative approach based on 3D CNNs and temporal action proposals has shown competitive results in~\cite{shou2016}  while \cite{zhao2017temporal} extends proposals in time and segments them in stages to evalute their ``completeness" based on structured pyramid pooling. 
Here we use GRU as recurrent units for temporal modeling of actions in our RecLNet.
Unlike previous work on temporal action localization, however, we address a much more challenging task of localizing actions in space and time.

\noindent{\bf Spatio-temporal detection.} Spatio-temporal action detection aims to find locations of actions in space and time.
The list of datasets for this task is limited: UCF101-24~-- a subset of UCF-101~\cite{soomro2012ucf101} used in the THUMOS challenge~\cite{THUMOS15}, the DALY dataset~\cite{weinzaepfel2016towards}.
Other datasets that we are aware of have either a very limited number of examples or consists of temporally trimmed videos. 
Some of the earlier works explore volumetric features and 3D sliding window detectors~\cite{ke2005efficient,laptev2007retrieving}.
More recent methods have extended ideas of object proposals in still images to action proposals in video~\cite{actiontubes,oneata2014spatio,gemert2015apt}.
The common strategy in~\cite{actiontubes,peng2016multi,saha2016deep,singh2017,Weinzaepfel_2015_ICCV,weinzaepfel2016towards} is to localize actions in each frame with per-frame action or person detectors and to link resulting bounding boxes into continuous tracks.
The temporal localization is then achieved by the temporal sliding windows~\cite{Weinzaepfel_2015_ICCV,weinzaepfel2016towards} or dynamic programming~\cite{peng2016multi,saha2016deep,saha2017amtnet,singh2017}. Instead of relying on per-frame detections, \citet{saha2017amtnet} regress pairs of successive frames
and \citet{hou2017tube} generate clip proposals from 3D feature maps. 
An approach of \citet{zolfaghari2017chained} uses a CNN to sequentially fuse human pose features in addition to the standard appearance and optical flow modalities.
Most recent method~\cite{kalogeiton17iccv} relies on SSD detector~\cite{liu2016ssd} adapted to spatio-temporal anchors and generates human tracks with tubelets linking. \cite{soomro2017unsupervised,yang2017common} investigate unsupervised spatio-temporal action localization but this is outside of the scope of our work.
In this paper, we propose a recurrent localization network (RecLNet) that both classifies and localizes actions within tracks supported by a thresholding and filtering method.
Our analysis shows that RecLNet provides significant gains due to accurate temporal localization and its complementarity to the recent method~\cite{kalogeiton17iccv} with more accurate spatial localization but approximate temporal localization. 
Our resulting approach outperforms the state of the art in spatio-temporal action detection~\cite{kalogeiton17iccv,peng2016hal,saha2016deep,singh2017,weinzaepfel2016towards} on two challenging benchmarks for this task.

\begin{figure*}
    \begin{center}
        \includegraphics[trim = 0mm 0mm 0mm 0mm, clip, scale=0.185]{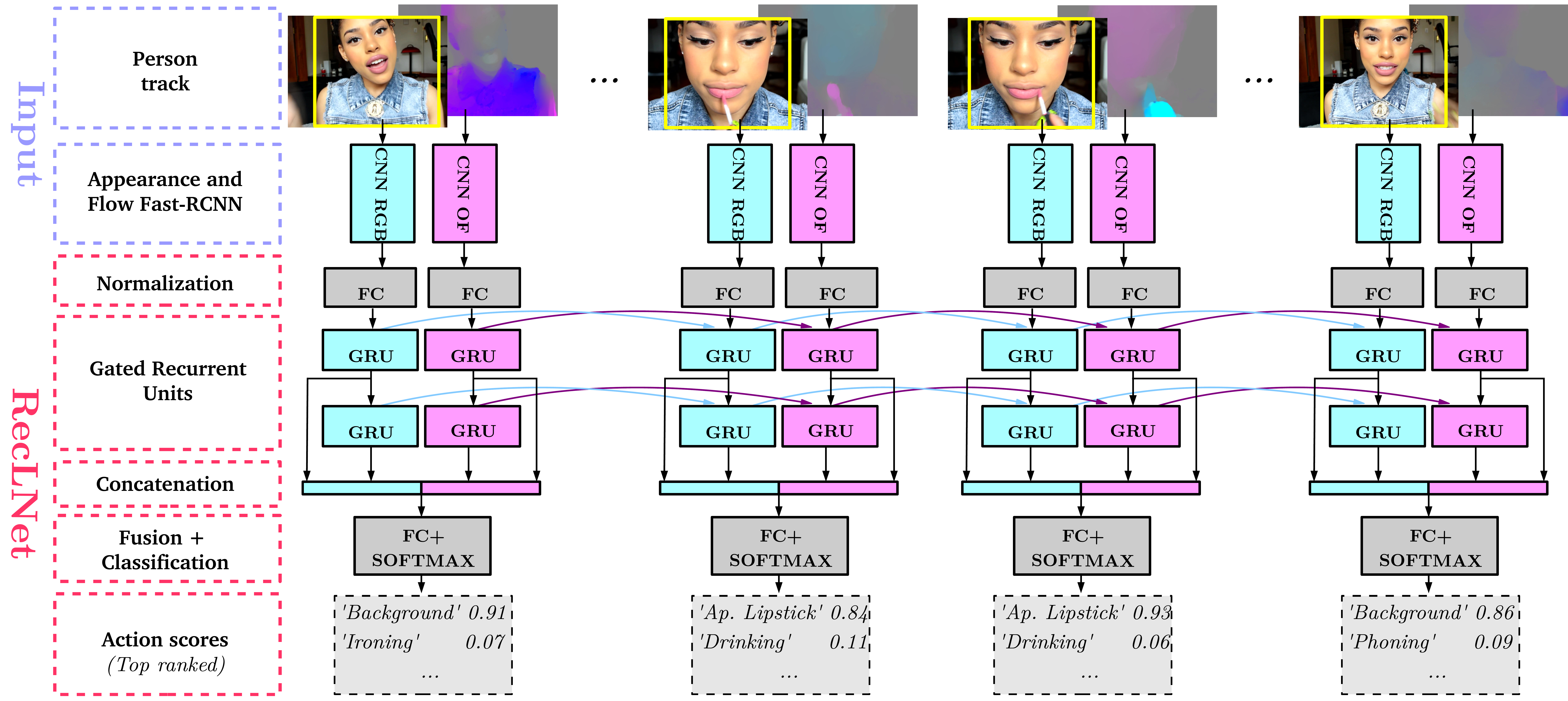} %
        
    \end{center}
    \caption{
Our RecLNet approach for spatio-temporal action localization. The input is a person track where for each spatially localized actor bounding-box, we extract appearance (RGB) and optical flow (OF) CNN features.  Each stream is normalized
by a fully-connected layer and fed into
a two-layer GRU. The outputs from both GRU levels and from both streams are concatenated then classified with a fully-connected layer combined with softmax scoring. Outputs are class probabilities for each frame. 
}
    \label{fig:pipeline}
\end{figure*}

\section{Action localization}
\label{sec:method}

This section presents our method for spatio-temporal action
localization. The overview of the method is illustrated in Figure~\ref{fig:pipeline}. A spatially localized person track (Figure~\ref{fig:pipeline}, row 1) is passed to appearance and optical flow feature extractors (Figure~\ref{fig:pipeline}, row 2).
These descriptors feed our localization network composed of 3 layers for each stream and 1 fusion layer.
In each stream, the first layer
(Figure~\ref{fig:pipeline}, row 3) normalizes
either appearance or flow features before sending them to a stack of two GRU layer
units. The GRU outputs from both stack levels and both streams are concatenated (Figure~\ref{fig:pipeline}, row 5)  and converted by a fully-connected layer
(Figure~\ref{fig:pipeline}, row 6) to action probabilities
(Figure~\ref{fig:pipeline}, last row).

In the following, we first briefly present the inputs of our method, namely the human
tracks and their associated features. Then, we introduce our
spatio-temporal action localization method based on the recurrent localization
network (RecLNet). Finally, we discuss how to post-process the action
detection scores in order to output final spatio-temporal human action
tubes.

\subsection{Person tracks}
\label{sec:tubenets}
We obtain human tracks as in~\cite{saha2016deep,weinzaepfel2016towards} by
first running an action or person detector in each video 
frame, and then linking detections into human tracks spanning the
entire video clip. 
Our method aims at segmenting the track in time to obtain temporal action
boundaries. For this purpose, we associate to each time frame its corresponding bounding-box in
the human track. This box is used as a pooling region to extract per-frame descriptors.
Such features are obtained by Fast-RCNN~\cite{Girshick_2015_ICCV} appearance and flow ROI-pooling.
The details on state-of-the-art human  
tracks and their associated features used in this work will be discussed
in Section~\ref{sec:datasets}.

\subsection{Temporal action localization}
\label{sec:rnn}

This section presents our model for action localization and its training procedure on the level of person tracks.
In our work, we choose to adopt a model with memory links, like in a recurrent neural network (RNN), which we call recurrent localization network (RecLNet), to 
temporally localize actions. RNNs have been demonstrated to successfully model sequential data especially for language tasks such as speech
recognition~\cite{dahl2012context}, machine translation~\cite{bahdanau2014neural} or image captioning~\cite{karpathy2015deep}.
Given this success, we believe that recurrent networks are well-suited for modeling temporal sequences of appearance and motion in person tracks. Our final RecLNet model is composed of gated recurrent units (GRU) described below.

\noindent{\bf The LSTM and GRU architectures.}
Features from a human track of length $T$ can be seen as an input sequence $\mathbf{x} = (x_1, ...,x_T)$
where for each $x_i$ we aim to provide action activation $h_i$, forming the output
$\mathbf{h} = (h_1, ...,h_T)$. To generate such output, we investigate two types of recurrent networks for our RecLNet, namely LSTM and GRU as defined below.

In the long short-term memory (LSTM) architecture~\cite{hochreiter1997long}, one memory cell and three `gates'
give LSTM the ability of discovering long-range temporal relationships by reducing the vanishing gradient problem compared to vanilla RNN. This is a useful property in our task since we need to handle particularly long human tracks.
The LSTM cell takes as input features~$x_t$ at time step~$t$ together with the output~$h_{t-1}$ at the previous time step~$t-1$ and operates as follows:
\begin{equation}
\begin{split}
f_t &= \sigma (W_f x_t + U_f h_{t-1} + b_f ) \\
i_t &= \sigma (W_i x_t + U_i h_{t-1} + b_i ) \\
o_t &= \sigma (W_o x_t + U_o h_{t-1} + b_o ) \\
c_t &= f_t \odot c_{t-1} + i_t \odot \tanh(W_c x_t + U_c h_{t-1} + b_c ) \\
h_t &= o_t \odot \tanh(c_t) \\
\end{split}
\end{equation}
where $\sigma$ is the sigmoid function and $f_t$, $i_t$, $o_t$, $c_t$ are the `forget gate',  `input gate', `output gate' and `memory cell',  respectively.
Matrices $W_\cdot$, $U_\cdot$ and vectors $b_\cdot$ denote the parameters of the cell.

The GRU~\cite{cho2014learning} cell differs from LSTM by the absence of the output gate and operates as follows:
\begin{equation}
\begin{split}
z_t &= \sigma (W_z x_t + U_z h_{t-1} + b_z ) \\
r_t &= \sigma (W_r x_t + U_r h_{t-1} + b_r ) \\
h_t &= z_t \odot h_{t-1}\\
&+ (1-z_t) \odot \tanh(W_h x_t + U_h (r_t \odot h_{t-1}) + b_h ) \\
\end{split}
\end{equation}
where $z_t$ and $r_t$ are the `update gate' and `reset gate', respectively.
The GRU cell is simpler, has less parameters and
has shown some improvement on video tasks as in~\cite{Tokmakov17}.
An empirical comparison of LSTM and GRU cells is given in Section~\ref{sec:experiments}.

\noindent{We define the \bf Recurrent Localization Network} (RecLNet)
as a multi-class recurrent network trained to classify actions against background.
As shown in Figure~\ref{fig:pipeline}, at each time step, appearance (RGB) and optical flow (OF) Fast-RCNN
networks take bounding boxes of human tracks as object proposals and extract
features. Each stream (RGB and OF) is processed independently by feeding Fast-RCNN outputs (FC7
layers) to our RecLNet which produces action scores at
each time step. Each stream of our RecLNet consists of a 
fully-connected layer that normalizes
the input features (appearance or flow) and a
stack of two GRU layers.
Note the second GRU layer takes as input the output of the first one
while their output is concatenated to a $2 \times M$-dimensional stream output (where $M$ is the memory size).
Finally the appearance and flow branch outputs are concatenated and a last fully-connected layer with softmax converts the
recurrent output to an action probability for all actions (and background).
The network is trained using the standard negative log-likelihood loss
w.r.t. all $C$ action classes and background boxes: 
\begin{equation}
\label{eq:loss}
L(W) = -\sum_{i=1}^N\sum_{c=1}^{C+1} \mathbbm{1}\{y_i=c\}\log P(y_i=c|x_i,W), \\
\end{equation}
with probabilities defined by softmax:
\begin{equation}
\label{eq:soft}
P(y_i=c|x_i,W) = \frac{\mathrm{e}^{f_c(x_i,W)}}{\sum_{c'=1}^{C+1} \mathrm{e}^{f_{c'}(x_i,W)}}. \\
\end{equation}
Here $f_c(x_i,W)$ is the output of the last fully-connected layer corresponding to class $c$, $x_i$ and $y_i$ denote features and labels at time step~$i$, symbol~$W$ denotes network parameters, $N$ is the number of boxes in the training set, $C$ is the number of action classes, $\mathbbm{1}\{.\}$ is the indicator function.

To train RecLNet, we set the ground-truth targets $y_i$ in the following way. We assign label $c$ to a frame bounding-box detection from a person track if it
overlaps more than 0.3 spatial IoU with one ground truth annotation from action label $c$, otherwise this input is considered as background.
Having low IoU threshold allows us to get more positives.

\noindent{\bf Appearance and optical flow fusion.}
Single-stream networks are first independently trained then
we consider three fusion methods to combine their appearance (RGB) and optical flow (OF) ouputs.
The \emph{average} simply averages both softmax RGB and OF network outputs.
Both next fusion methods are trained on top of the two stream networks (their weights remaining fixed).
The \emph{gating layer} learns per-class weights to multiply both network outputs before their original softmax layer and applies softmax after summing weighted class outputs.
The third method, \emph{fusion layer}, trains a fully-connected classification layer (followed by softmax)
on top of the concatenated memory units outputs from each stream network (concatenation layer in Figure~\ref{fig:pipeline}).

\vspace{-4mm}
\subsection{Post-processing of human action tracks}
\label{sec:postproc}
RecLNet output represents action scores at each time step of a human track.
While being spatially localized, these scored tracks have to be segmented in time in order to produce the final spatio-temporal detections.
For the track ranking, it is also necessary to score each of the final detections.

\noindent{In this section, we then describe our method to get the final spatio-temporal
detections and their associated score.
Temporal localization is performed within each track.}

\noindent{\bf Threshold.}
Let $a$ be the action to segment and $\tau$ a human track spanning from frame $1$ to $T$ of scores
$\mathbf{s_\tau} = (s_1^a, ...,s_T^a)$ with $s^a_t$ the score associated to the person box at time $t$.
The goal of temporal segmentation is to extract from $\tau$ one or several sub-tracks $\nu$ (of time interval $[t_1^\nu,t_2^\nu] \subseteq [1,T]$)
as final spatio-temporal detections.

\noindent For this purpose, we first get smoother scores $\hat{s}^a_i$ by applying a median window filtering on the $s^a_i$.
Then, we temporally segment the track by selecting consecutive boxes
with scores above a certain threshold $\theta$, while others are rejected.
More formally, considering that consecutive boxes $(b_{t_0}, ..., b_{t-1})$ from time $t_0$ to $t-1$ have already been added to the sub-track $\nu$, the next box $b_t$ is added to $\nu$ if $\hat{s}^a_t \geq \theta$. Otherwise, $\nu$ ends and is returned as final detection.
This method allows to break the initial track into several sub-track candidates of arbitrary lengths
and is then able to capture several repetitive action instances  
happening on the same human track (like \emph{drinking}, \emph{applying make up}, \emph{playing harmonica}, see Figure~\ref{scoreblobs1} and Figure~\ref{scoreblobs2}).
Here, model outputs must be smooth in order to get accurate action temporal boundaries.
The recurrent units (like GRU, LSTM) are
then well-suited for this localization method while appearance and flow CNNs output is generally noisier.
Our temporal localization technique is referred to
as \emph{threshold}.

\noindent{\bf Temporally segmented track scoring.}
In order to rank detections and perform the average-precision (AP) evaluation, we need to set a score $r_\nu$ for each final spatio-temporal detection $\nu$.
Following, e.g.,~\cite{saha2016deep}, we define $r_\nu$ as the average
of the top 40 action scores $s^a_i$ contained in $\nu$.
The same scores are also used for non-maximum-suppression (NMS)
of spatio-temporal detection candidates
based on their scores and overlap. 
For NMS, we use spatio-temporal
Intersection-over-Union\footnote{The spatio-temporal IoU between two
  tracks is defined as a product of temporal IoU between the time
  segments of the tracks and average spatial IoU on the frames where
  both tracks are present.} as overlap criterion.

\subsection{Implementation details}

\noindent{\bf RecLNet parameters.} The FC7 output of each Fast-RCNN is $4096$-dimensional and the first fully-connected layers convert each stream to a $1024$-dimensional vector. The memory size $M$ is equal to 256 and the last fully-connected layer input is $1024$-dimensional ($2 \times 2 \times M$, i.e. the memory stack from both GRU layers of both streams).

\noindent{\bf Training.}
Both appearance and flow branches of RecLNet are separately trained using the Adam optimizer~\cite{kingma2014adam}
with a weight decay set to $5.10^{-4}$ to avoid overfitting.
Note that to train the single-stream networks, we halve the input dimension of the last fully-connect layer of RecLNet (concatenation layer in Figure~\ref{fig:pipeline}).
Training batches contain $100$ different tracks of temporal length $20$.
Backpropagation through time (BPTT) is then performed every $20$ time steps.

\noindent{\bf Detection.}  In all experiments, the NMS overlap threshold is set 0.2, the action localization \emph{threshold} $\theta$ is set
to $0.1$ and the median window size is 25.

\noindent{\bf Optical flow.} To obtain the optical flow data, we
compute horizontal and vertical flow for each consecutive pair of
frames using the approach
of~\cite{brox2004high}. Following~\cite{actiontubes,Weinzaepfel_2015_ICCV},
flow maps are saved as 3-channels images corresponding to optical flow
in x and y direction and its magnitude with all the values restricted to the
interval $[0,255]$.

\section{Experimental setup}
\label{sec:datasets}

This section describes UCF101-24~\cite{soomro2012ucf101} and DALY~\cite{weinzaepfel2016towards} datasets used for evaluation of our method.
For both datasets, we provide experimental details on the use of ground truth annotation and data pre-processing.

\subsection{UCF101-24}
\label{sec:datasets_ucf}

The original version of the UCF-101 dataset~\cite{soomro2012ucf101} is designed for action classification and contains 13321 videos for 101 action classes.
The task of spatio-temporal action localization is defined on a subset of 24 action classes (selected by~\cite{THUMOS15}) in 3207 videos.
We refer to this subset as ``UCF101-24".
Each instance of an action is manually annotated by a person track with the temporal interval corresponding to the interval of an action.
In our training and testing, we use recently corrected ground truth tracks~\cite{saha2016deep}.\footnote{\url{https://github.com/gurkirt/corrected-UCF101-Annots}}
Each UCF-101 video contains actions of a single class.

\noindent{\bf Short vs. long classes.}
While some of the UCF101-24 action classes are short ('basketball dunk' or 'tennis swing') other continuous actions ('biking' or 'rope climbing') typically last for the full duration of the video.
To better evaluate the temporal localization on UCF101-24, we define a subset with short action classes that on average last less than a half of the video length.
This subset contains six actions (\emph{Basketball}, \emph{Basketball Dunk}, \emph{Cricket Bowling}, \emph{Salsa Spin}, \emph{Tennis Swing} and \emph{Volleyball Spiking}) and we call them ``Short classes''.
In Section~\ref{sec:experiments}, we evaluate localization for all 24 action classes and for Short classes separately.

\noindent{\bf Performance evaluation.}
To evaluate the detection performance, we use the standard spatio-temporal IoU criterion defined by the UCF101-24 benchmark.
The detected action tube is considered to be correct if and only if its intersection with the ground-truth tube is above the criterion threshold and if both tubes belong to the same action class.
Duplicate detections are considered as false positives and are penalized by the standard precision-recall measure.
The overall performance on the UCF101-24 dataset is compared in terms of mean average precision (mAP) value.

\noindent{\bf Action-specific human tracks.}
To enable direct comparison of our method with~\cite{saha2016deep}, for UCF101-24 experiments we use the same person tracks as in~\cite{saha2016deep}.
These tracks are obtained by linking per-frame action detections with dynamic programming (DP).
Action detections are obtained with the Fast-RCNN method~\cite{Girshick_2015_ICCV} trained with appearance and flow input for the task of spatial action localization.
As our model performs its own temporal localization, we do not run the temporal segmentation of~\cite{saha2016deep} (second DP pass) and keep 5 action proposals per action covering the whole video.
Per-frame input features for our RecLNet are obtained from the same Fast-RCNN detector used for track detection.

\subsection{DALY}
\label{sec:datasets_daly}
DALY~\cite{weinzaepfel2016towards} is a recent large-scale dataset for action localization containing 510 videos (31 hours) of 10 different daily activities such as 'brushing teeth', 'drinking' or 'cleaning windows'. There is only one split containing 31 train and 20 test videos per class. The average length of the videos is 3min 45s.
In contrast to UCF101-24, all actions are short w.r.t.~the full video length, making the task of temporal action localization more challenging.
DALY may contain multiple action classes in the same video.
The DALY dataset provides ground-truth temporal boundaries for all the action instances and the spatial annotation (bounding boxes) for few keyframes of each instance.

\noindent{\bf Human tracks and associated features.}
To enable direct comparison with~\cite{weinzaepfel2016towards} we use tracks provided by~\cite{weinzaepfel2016towards}.
An accurate Faster-RCNN~\cite{ren2015faster} person detector is trained on the large MPII human pose dataset~\cite{andriluka14cvpr}.
The tracks are obtained by linking human detections with a tracking-by-detection approach.
Similarly to UCF101-24, the appearance and flow features are obtained with a Fast-RCNN detector
trained to detect actions on the annotated DALY frames. The action scores from this detector
will be used as a baseline in Section~\ref{sec:experiments}
and referred to as \emph{CNN RGB+OF}.
Following~\cite{Mathias2014Eccv}, we compensate the annotation bias by adapting human box sizes to action annotation. To achieve this, we train a linear regression of bounding boxes.
For each DALY annotated keyframe from the training set we associate its most overlapping bounding-box from the human tracks (with at least 0.5 IoU) to train the linear model.

\noindent{\bf DALY tracks labeling.}
To compensate for the sparse action annotation in DALY, we extend ground truth to all frames of the actions with automatic tracking. We use the Siamese Fully-Convolutional Network for online non-class-specific tracking method\footnote{\url{https://github.com/bertinetto/siamese-fc}}~\cite{bertinetto2016fcsn} initialized on all ground-truth keyframes. For a given action instance, we aggregate all tracks at each frame by a median bounding box. To assess the quality of generated ground-truth tracks, we computed the track proposals recall ($92.2\%$ at $0.3$) within the action time interval. Also, the IoU overlap between ground-truth tracks and annotated keyframes is greater than $0.8$ (resp. $0.5$) for $91.4\%$ (resp. $99.0\%$) of keyframes. %

\section{Experiments}
\label{sec:experiments}

In this section, we first evaluate the impact of the \emph{threshold} temporal localization, recurrent architecture and the fusion method (Section~\ref{sec:exp_imp}). We then  
evaluate the potential gain due to \emph{action classification} (Section~\ref{sec:explain_gain}) followed by an extensive analysis on \emph{temporal localization} (Section~\ref{sec:explain_gain_temp}). Next, we show an improvement if using I3D and temporal person tracks (Section~\ref{sec:repbaseline}).
We compare our RecLNet to the state of the art on UCF101-24 and DALY datasets (Section~\ref{sec:comp}).
Section~\ref{sec:qualitative} concludes this experimental part by presenting qualitative results.

\subsection{Impact of localization method, recurrent architecture and fusion} 
\label{sec:exp_imp}

\begin{table*}[t]
\setlength{\tabcolsep}{3mm}
\def\arraystretch{1.2}
    \begin{center}
       \begin{tabular}{lccccccc}
        &  \multicolumn{3}{c}{UCF101-24} & \multicolumn{3}{c}{DALY} \\
         Method & OF & RGB & RGB+OF & OF & RGB & RGB+OF \\
        \hline \hline 
        CNN~\cite{saha2016deep} & 59.3 & 55.2 & 60.4 & - & - &- \\
        CNN & - & - & - & 9.9 & 11.9 & 13.4 \\
        \emph{FC}   & 60.7& 57.7&64.0 & 11.5 & 13.6 & 16.1 \\
        \hline
        LSTM    & 65.0 & 58.7 & 66.5 & 13.1 & 11.9 & 16.2  \\
        GRU    & \textbf{67.0} & \textbf{59.5}&\textbf{67.1} & \textbf{14.4} & \textbf{14.2} & \textbf{17.4} \\
    \end{tabular}
    \end{center}
    \caption{Performance on UCF101-24 and DALY for different recurrent architectures
    (\emph{LSTM} and \emph{GRU}) and baselines with no temporal connections (\emph{CNN} and \emph{FC}).
    Models are evaluated with flow (OF) and/or appearance (RGB) features as input (RGB+OF averages both stream outputs) for spatio-temporal localization at IoU = $0.3$ (mAP).
        \label{tbl:hyperparamArch}}
\end{table*}

\noindent{\bf Localization method.}

In~\cite{saha2016deep}, temporal localization is performed using the Viterbi algorithm on top of action tracks spanning the whole video.
To temporally trim these tracks, one binary label (action vs. background) is associated to each bounding-box by maximizing an energy function where unary potentials represent
box action scores while pairwise potentials control the final action length (or track smoothness).
This smoothness score is weighted per class and then might over control the track length towards the action durations seen on the training set
while, in our method, median filtering encodes smoothness more explicitly and suffers less from the dataset biases described in~\ref{sec:datasets_ucf}.

In Table~\ref{tbl:hyperparamLoc}, we refer to the track action scores from~\cite{saha2016deep}
as \emph{CNN RGB+OF}, since they come from two Fast-RCNN where appearance and optical flow outputs have been fused. 
We recall that to enable a direct comparison with~\cite{saha2016deep}
we are using the same human tracks as~\cite{saha2016deep}
but score them differently (see Section~\ref{sec:datasets} for
details). Here, we compare the Viterbi algorithm originally
used in~\cite{saha2016deep} for temporal localization and our \emph{threshold} method on original scores (\emph{CNN RGB+OF}).
Interestingly we observe that our \emph{threshold} technique improves over
their original results by $+4.9\%$ and conclude that thresholding combined with median filtering works better
than this typically used localization method while being simpler.
The next paragraph shows the impact of differently scoring, among others, tracks from~\cite{saha2016deep}.
In the following, only our \emph{threshold} temporal localization is used. 

\begin{table}[h]
\setlength{\tabcolsep}{3mm}
\def\arraystretch{1.2}
    \begin{center}
       \begin{tabular}{lcc}
        Method & Localization & UCF101-24  \\
        \hline \hline 
        CNN RGB+OF &  Viterbi~\cite{saha2016deep}  & 55.5   \\
        CNN RGB+OF & threshold & 60.4  \\
    \end{tabular}
    \end{center}
    \caption{Localization method analysis on UCF101-24.
     We compare the temporal localization of~\cite{saha2016deep}
     which uses \emph{Viterbi} algorithm to our \emph{threshold} method.
     We apply the two methods on original detection scores from~\cite{saha2016deep} (\emph{CNN RGB+OF}).
     Evaluation is spatio-temporal action localization at IoU 0.3 (mAP).
        \label{tbl:hyperparamLoc}}
\end{table}

\noindent{\bf Recurrent architecture.} 
Table~\ref{tbl:hyperparamArch} compares models based on LSTM and GRU units (see Section~\ref{sec:method} for details).
As a baseline we use a model with no temporal connections but similar architecture:
the recurrent units are replaced by a stack of
2 fully-connected layers with non-linearity and has the same number of parameters as the GRU unit. This additional fully-connected classifier
is referred to as \emph{FC}. For UCF101-24 evaluation, we again report original track scores (\emph{CNN~\cite{saha2016deep}} as in Table~\ref{tbl:hyperparamLoc}) while for DALY, \emph{CNN} represents the Fast-RCNN scores similarly retrained as authors~\cite{weinzaepfel2016towards} on their human tracks (see Section~\ref{sec:datasets_daly} for details). Input modalities are either, optical flow features (\emph{OF}),
appearance features (\emph{RGB}) or both (\emph{RGB+OF}) where action scores from the two stream outputs are averaged in this case. Spatio-temporal action localization is evaluated on UCF101-24 and DALY datasets at spatio-temporal IoU of 0.3.

We first observe that training an additional classifier (\emph{FC}) is better than directly taking Fast-RCNN outputs (\emph{CNN})
and improves performance by $+3.6\%$ and $+2.7\%$ on UCF101-24 and DALY respectively
when both appearance and optical flow streams are used (\emph{RGB+OF}).
Also, non-recurrent baselines (\emph{CNN} and \emph{FC}) perform worse than the recurrent variants (\emph{LSTM} and \emph{GRU}).
When using both \emph{RGB+OF} as input, the gain due to recurrence is $+6.7\%$ and $+4.0\%$ on UCF101-24 and DALY respectively when comparing \emph{GRU} to \emph{CNN}
and $+3.1\%$ and $+1.3\%$ when comparing \emph{GRU} to \emph{FC}.
Interestingly, recurrent unit improvement is larger on optical flow $+6.3\%$ on UCF101-24 and $+2.9\%$ on DALY
when comparing $GRU$ to $FC$.
This shows that temporal memory links are beneficial for better spatio-temporal action localization performance.
We also note, as this is often the case when working with videos (e.g.~\cite{Tokmakov17}),
that  \emph{GRU} outperforms \emph{LSTM} while being
a simpler model.

This experiment has shown \emph{GRU} memory unit achieves the best action localization accuracy.
In the following, \emph{GRU} will then be used in our recurrent model for
determining the action temporal extent in all experiments.

\noindent{\bf Fusion method.} Table~\ref{tbl:hyperparamFusion} explores the different fusion strategies described in Section~\ref{sec:method}
to combine appearance (RGB) and flow (OF) features in our localization model (with \emph{GRU} layer which
was validated in the previous experiment). We first note that all fusion methods improve
action localization results on both UCF101-24 and DALY. However, the simple averaging method is
not able to capture feature complementarity between appearance and flow features especially on UCF101-24
where it gets only $+0.1\%$ improvement compared to the best performing OF features. The \emph{gating layer}
is able to take advantage of features combination by substantially improving on both datasets. Finally, the \emph{fusion layer}
better captures features complementarity and improves action localization mAP on UCF101-24 and DALY respectively by $+2\%$ and $+5.3\%$ compared to the single flow features and by $+9.5\%$ and $+5.5\%$ compared to appearance features.

In the following, our final recurrent model uses the \emph{fusion layer} and is referred to as \emph{RecLNet} in all experiments. 

Finally, when using our \emph{threshold} method, re-scoring the tracks with our RecLNet instead of taking Fast-RCNN outputs (\emph{CNN}) as in~\cite{saha2016deep,weinzaepfel2016towards}
improves spatio-temporal action localization results by $+8.6\%$ and $+4.0\%$ on UCF101-24 and DALY respectively (from Table~\ref{tbl:hyperparamArch}).
The improvement is even larger ($+13.5\%$) when comparing RecLNet accuracy ($69.0\%$) to the original result using scores and temporal localization from~\cite{saha2016deep} ($55.5\%$ on UCF101-24 from Table~\ref{tbl:hyperparamLoc}).
 
Given that we are using the same tracks, this first gives an insight that
our method improvement compared to~\cite{saha2016deep,weinzaepfel2016towards} might be due to \emph{temporal} localization. This question is further studied in the next sections.

\begin{table}[H]
\setlength{\tabcolsep}{3mm}
\def\arraystretch{1.2}
    \begin{center}
       \begin{tabular}{lccc}
        Features & Fusion  & UCF101-24 & DALY \\
        \hline \hline 
   	OF &-& 67.0 & 14.4 \\
   	RGB &-& 59.5 & 14.2 \\
        \hline
        RGB+OF    & \emph{average} & 67.1 & 17.4 \\
        RGB+OF    & \emph{gat. layer} & 69.0 & 18.1 \\
        RGB+OF    & \emph{fusion layer} & \textbf{69.0} & \textbf{19.7} \\
    \end{tabular}
    \end{center}
    \caption{Performance on UCF101-24 and DALY for different features and fusion strategies \emph{average}, \emph{gating layer} and \emph{fusion layer} (see Section~\ref{sec:method} for details).
    Models with \emph{GRU} memory units take optical flow (OF) and/or appearance (RGB) features as input
    and are evaluated for spatio-temporal localization at IoU = $0.3$ (mAP).
        \label{tbl:hyperparamFusion}}
\end{table}

\vspace{-15mm}
\subsection{Analyzing the performance gain: \emph{action classification}}
\label{sec:explain_gain}
Spatio-temporal action localization is composed of \emph{spatial localization}, \emph{action classification} and \emph{temporal localization}.
In this section we focus on action classification and analyze its performance given person tracks and pre-defined temporal action boundaries.

\noindent{\bf Evaluating action classification.}
Here, we do not require temporal localization by restricting the track to the ground-truth time interval
in order to evaluate the performance gain due to action classification.
As described in Section~\ref{sec:method}, the track scoring is obtained
with the average over the top 40 action scores $s^a_i$ in the trimmed interval. 
Table~\ref{tbl:noTemporalLocalization} shows our method (\emph{RecLNet}) increases action classification
by approximately $+2\%$ compared to scores from~\cite{saha2016deep} (\emph{CNN RGB+OF}) when
evaluated on UCF101-24 classes. This improvement is moderate compared to the $+13.5\%$ spatio-temporal localization boost of Section~\ref{sec:exp_imp}.

\begin{table}[h]
\setlength{\tabcolsep}{2mm}
\def\arraystretch{1.2}
    \begin{center}
        \begin{tabular}{lcccc}
          {IoU}  && 0.2 & 0.3 & 0.4 \\
            \hline
            \hline
            CNN RGB+OF~\cite{saha2016deep}~~~~~~~~  && 84.3 & 81.6	& 75.6 \\
            RecLNet && 86.9 &	83.7 &	77.6 \\
        \end{tabular}
    \end{center}
    \caption{Performance on UCF101-24 for clips trimmed to the ground-truth temporal interval (mAP at IoU 0.2, 0.3 and 0.4). 
        \label{tbl:noTemporalLocalization}}
\end{table}

Table~\ref{tbl:DALYnoTemporalLocalization} shows the same experiments on DALY. We observe that our \emph{RecLNet} method performs slightly worse (from $-0.1\%$ to $-0.9\%$) compared to the \emph{CNN RGB+OF} baseline (see Section~\ref{sec:datasets_daly} for details). This might be due
to DALY track labeling which, contrary to UCF101-24, is obtained by automatic tracking of ground-truth sparse annotations (see Section~\ref{sec:datasets_daly}) which introduces some noise and can slightly affect action classification. We, therefore, conclude that the spatio-temporal action localization gain of our method over the baseline, demonstrated in Section~\ref{sec:exp_imp}, cannot come from action classification.

\begin{table}[h]
\setlength{\tabcolsep}{2mm}
\def\arraystretch{1.2}
    \begin{center}
        \begin{tabular}{lcccc}
          {IoU}  && 0.2 & 0.3 & 0.4 \\
            \hline
            \hline
            CNN RGB+OF  && 65.5 & 64.8 & 63.6 \\
            RecLNet 			&& 65.4 &64.5 &	62.7 \\
        \end{tabular}
    \end{center}
    \caption{Performance on DALY for clips trimmed to the ground-truth temporal interval (mAP at IoU 0.2, 0.3 and 0.4). 
        \label{tbl:DALYnoTemporalLocalization}}
\end{table}

Overall, this experiment confirms results in the 
literature~\cite{Ng_2015_CVPR} that recurrent units (RNN, LSTM, GRU) do not really improve the
performance for action classification. The spatial localization (the person boxes positions) being fixed, the gain of our method for spatio-temporal action localization has to come from better temporal localization.
This observation is analyzed in detail in Section~\ref{sec:explain_gain_temp}.

\begin{table*}
    \newcommand\sih{\small} %
    \newcommand\sihnum{\scriptsize} %
    \setlength{\tabcolsep}{0.07cm}
    \centering 
    \begin{tabular}{lccccccCccccccccccccccccccC}    
    \multicolumn{27}{r}{
    \includegraphics[trim = 40mm 42mm 14mm 0mm, clip, width=147mm,height=13mm]{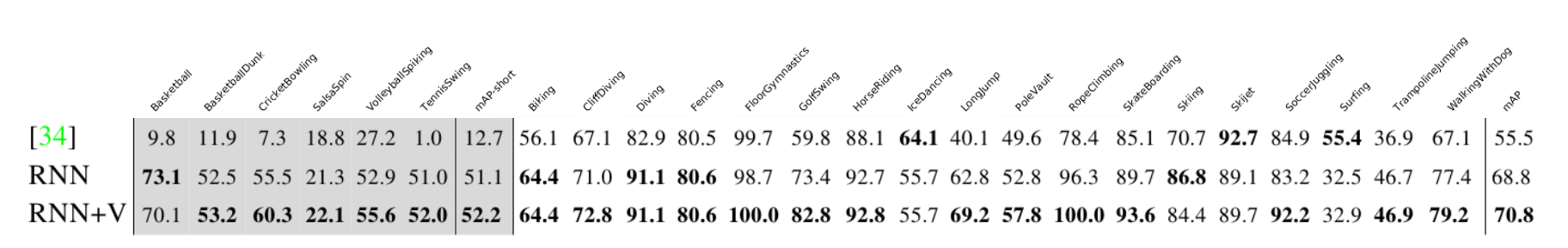} %
    }
	\\
        \hline
        {\sih \cite{saha2016deep}}
        &{\sihnum  9.8}
        &{\sihnum  11.9}
        &{\sihnum  7.3}
        &{\sihnum  18.8}
        &{\sihnum  1.0}
        &{\sihnum  27.2}
        &{\sihnum 12.7}
        
        &{\sihnum  56.1}
        &{\sihnum  67.1}
        &{\sihnum  82.9}
        &{\sihnum  80.5}
        &{\sihnum  \textbf{99.7}}
        &{\sihnum  59.8}
        &{\sihnum  88.1}
        &{\sihnum  \textbf{64.1}}
        &{\sihnum  40.1}
        &{\sihnum  49.6}
        &{\sihnum  78.4}
        &{\sihnum  85.1}
        &{\sihnum  70.7}
        &{\sihnum  \textbf{92.7}}
        &{\sihnum  84.9}
        &{\sihnum  55.4}
        &{\sihnum  36.9}
        &{\sihnum  67.1}
        &{\sihnum 55.5}
        \\
        {\sih  RecLNet}
        &{\sihnum \textbf{69.4}}
        &{\sihnum 38.8}
        &{\sihnum \textbf{50.8}}   
        &{\sihnum \textbf{25.4}}  
        &{\sihnum \textbf{26.5}} 
        &{\sihnum \textbf{53.0}}
        &{\sihnum \textbf{44.0}}
        
        &{\sihnum  65.0}
        &{\sihnum  75.6}
        &{\sihnum  \textbf{93.4}}
        &{\sihnum  80.6}
        &{\sihnum  99.6}
        &{\sihnum  70.3}
        &{\sihnum  93.0}
        &{\sihnum  55.4}
        &{\sihnum  65.6}
        &{\sihnum  56.9}
        &{\sihnum  \textbf{99.3}}
        &{\sihnum  91.6}
        &{\sihnum  \textbf{84.8}}
        &{\sihnum  89.3}
        &{\sihnum  \textbf{92.3}}
        &{\sihnum  51.6}
        &{\sihnum  47.5}
        &{\sihnum \textbf{80.5}}
        &{\sihnum \textbf{69.0}}
        \\
        \hline
         {\sih\cite{kalogeiton17iccv}}
        &{\sihnum  18.0}
        &{\sihnum  \textbf{44.1}}
        &{\sihnum  23.5}
        &{\sihnum  20.7}
        &{\sihnum  1.0}
        &{\sihnum  39.1}
        &{\sihnum  24.4}
        
        &{\sihnum \textbf{72.0}}
        &{\sihnum \textbf{82.0}}
        &{\sihnum 84.8}
        &{\sihnum \textbf{81.3}}
        &{\sihnum 99.2}
        &{\sihnum \textbf{76.7}}
        &{\sihnum 94.6}
        &{\sihnum 60.8} 
        &{\sihnum \textbf{82.5}}
        &{\sihnum \textbf{90.4}}
        &{\sihnum 92.4}
        &{\sihnum \textbf{92.0}}
        &{\sihnum 84.0}
        &{\sihnum 80.4}
        &{\sihnum 91.4}
        &{\sihnum \textbf{65.9}}
        &{\sihnum \textbf{62.3}}
        &{\sihnum 76.8}
        &{\sihnum 67.3}      
    \end{tabular}
\vspace{3mm}
    \caption{Per-class AP for IoU 0.3 on UCF101-24 comparing our RecLNet
      to state of the art~\cite{saha2016deep,kalogeiton17iccv}. We report mean AP for
      short classes, \emph{mAP-short}, and all classes,
      \emph{mAP}. The 6 short classes are reported on the left (before \emph{mAP-short} column).
      \label{tbl:perClassSota}}
\end{table*}

\subsection{Analyzing the performance gain: \emph{temporal localization}}
\label{sec:explain_gain_temp}
In this section, we show the main gain of RecLNet comes from better temporal localization and demonstrate its
complementarity with the state of the art~\cite{kalogeiton17iccv}.
We first motivate this study by explaining the UCF101-24 bias, then a per-class analysis compares our RecLNet to~\cite{kalogeiton17iccv} and to the method we build on~\cite{saha2016deep}. Finally, we investigate the potential room for improvement.

\noindent{\bf UCF101-24 bias.}
As described in Section~\ref{sec:datasets_ucf}, UCF101-24 contains only six short action classes lasting 
less than half of the video duration, while 17 actions ($74\%$ of the dataset classes) span at least $70\%$ of the
video length in which 11 of them ($48\%$ of the dataset classes) span even more than $90\%$ of the video length.
UCF101-24 results are then biased toward long actions which do not require  temporal localization.
Indeed, tracks spanning the whole video already achieves good temporal localization for these classes (note that such detections would totally fail on the DALY dataset).
To avoid this bias and to focus on challenging cases of temporal localization, we next compare our method to~\cite{kalogeiton17iccv,saha2016deep} on the subset of short classes along with whole dataset.

\vspace{10mm}
\smallskip
\noindent{\bf Per-class performance on UCF101-24.}
Table~\ref{tbl:perClassSota} compares the per-class results of our approach 
with~\cite{saha2016deep} since we are using their human tracks and
with the best performing state-of-the-art method~\cite{kalogeiton17iccv} (see Section~\ref{sec:comp}
for comparison with the state of the art). 
We report mAP for short classes only (\emph{mAP-short}) and also for all classes (\emph{mAP}).
We outperform~\cite{saha2016deep} by $+31.3\%$ on average on short
classes and by $+59.6\%$ on ``Basketball''.  
As we have already seen previously, the
overall improvement is $+13.5\%$. We can observe that for 
some ``long'' actions the performance drops slightly. Similarly,
we outperform the state of the art~\cite{kalogeiton17iccv} by only $+1.7\%$ overall
while we reach a large improvement of around $+20\%$ on short classes.

\begin{table}[H]
\setlength{\tabcolsep}{1.2mm}
\def\arraystretch{1.2}
    \begin{center}
        \begin{tabular}{lcccccccc}
             &&  \multicolumn{3}{c}{All classes} &&  \multicolumn{3}{c}{Short classes}\\       
            {\small IoU}  && 0.3 & 0.5 & 0.75 && 0.3 & 0.5 & 0.75 \\
            \hline            \hline
            \cite{kalogeiton17iccv}  	&& 67.3 & \textbf{51.4} &  \textbf{22.7} && 24.4 & 2.6  & \textbf{0.0}  \\
            RecLNet 					&& \textbf{69.0} & 46.5 & 10.3  && \textbf{44.0} & \textbf{6.4} & \textbf{0.0}
        \end{tabular}
    \end{center}
    \caption{Performance on UCF101-24 when differentiating short classes from others.
      Spatio-temporal action localization (mAP) is evaluated at IoU 0.3, 0.5 and 0.75).
     \label{tbl:shortLongIoU}}
\end{table}

Table~\ref{tbl:shortLongIoU} now compares our model to the best performing method~\cite{kalogeiton17iccv} at higher IoU.
We observe that even if short classes get extremely difficult to detect at highest IoU ($0\%$ mAP at IoU 0.75), at IoU 0.5, our model still
outperforms the state of the art by $+3.8\%$ on short classes while loosing $-4.9\%$ overall
(Section~\ref{sec:comp} studies this latter result when comparing to the state of the art).

These experiences on per-class performance show
that our model is able to improve~\cite{saha2016deep} by producing more accurate temporal action localization.
Also, while~\cite{kalogeiton17iccv} takes advantage from strong features and spatially accurate person boxes to get
excellent accuracy on long classes (e.g. compared to~\cite{saha2016deep}), its performance on short classes suffers from approximate temporal localization. This demonstrates the potential complementarity between our RecLNet and the current best performing method~\cite{kalogeiton17iccv}.
The room for improvement of RecLNet will be studied in the next section.

\begin{table}[H]
\setlength{\tabcolsep}{2mm}
\def\arraystretch{1.2}
    \begin{center}
        \begin{tabular}{cccccc}
                 Correct 
                 & Correct 
                 & Correct
                 && \multicolumn{2}{c}{class mAP @0.75} \\
                 temporal loc. & spatial loc. & class &&  short & all \\
            		\hline
            - & - & - &&0.0& 10.3 \\
            \checkmark & - & - & &7.7&15.8 \\
            - & \checkmark & - &&\textbf{11.1}& \textbf{54.0} \\
            - & - & \checkmark &&0.0& 17.0 \\
            \hline
            \checkmark & - & \checkmark &&11.1& 23.8 \\
            - & \checkmark & \checkmark &&17.2& 60.8 \\
            \checkmark & \checkmark & - &&41.7&  70.5\\
            \checkmark & \checkmark & \checkmark &&43.2& 73.9 \\
            \hline
        \end{tabular}
    \end{center}
    \caption{RecLNet performance on UCF101-24 short and all classes when considering different components to be correct.
      Spatio-temporal action localization (mAP) is evaluated at IoU 0.75). \textbf{Spatial} localization
      is the largest room for improvement of our method.
     \label{tbl:correctness}}
\end{table}

\smallskip
\noindent{\bf Toward action localization improvement.}
This paragraph analyzes how to improve the spatio-temporal action localization performance of our model especially when evaluating at very high IoU (0.75).
To distinguish the potential improvements, we can consider \emph{action classification}, \emph{spatial localization} and/or \emph{temporal localization} as being correct. We proceed as follow.
When we suppose perfect \emph{action classification}, we set final spatio-temporal detection score $r_\nu$ (see Section~\ref{sec:postproc}) to 0 for all false positives (note this is equivalent to the recall).
Let $o_s$ (resp. $o_t$) the spatial IoU overlap of a final detection with its ground-truth. 
Thereby, when considering perfect \emph{spatial} (resp. \emph{temporal}) localization, if $o_s$ (resp. $o_t$) is greater than 0.3, it is then set to 1 in the spatio-temporal IoU computation.
We choose 0.3 as it seems fair enough 
1) not to get a track completely shifted in time ($o_t \simeq 0$) for which the spatial IoU $o_s$ would have been computed on only one frame or a few
and 2) to ensure that the track at least approximately spatially ``follows" ($o_s \neq 0$) the person. 
Table~\ref{tbl:correctness} presents the combinations of these assumptions.
First, assuming correct \emph{temporal localization} or \emph{action classification} only improves our model performance by
respectively $+5.5\%$ and
$+6.7\%$. However, the \emph{spatial} assumption is by far the best room for improvement of our model ($+43.7\%$) and achieves $54.0\%$ accuracy.
It also shows our method gets $0\%$ accuracy on short classes mostly because of inaccurate human track bounding-boxes as the spatial assumption  improves it to $11.1\%$.
Of course, combining several assumptions further improve the performance (note that last line does not reach $100\%$ since the track recall is not $100\%$ and that the above ``0.3 criterion" eliminates some track candidates). However, we observe that combining correct \emph{temporal localization} and \emph{action classification} ($23.8\%$) is still far from the single \emph{spatial localization} assumption ($54.0\%$). This experiment confirms that improving spatial human detection is the direction to take further enhancements and demonstrates that temporal localization is already a strong component of RecLNet.

This section has shown that the UCF101-24 bias is not in favour of our model while the short classes are by far the hardest to localize and then are
source of potentially large improvement for current action localization methods.
Also, we observed that our model already benefits from accurate temporal localization
while its considerable room for improvement is the spatial localization component
(the temporal localization being the one with the less potential for RecLNet). Consequently, the state of the art~\cite{kalogeiton17iccv}, which mostly relies on spatially more accurate human tracks and better features that the ones we build on~\cite{saha2016deep} (as previously compared in Table~\ref{tbl:correctness}), and our model,
which greatly improves temporal localization, are definitely complementary.

\subsection{Improved tracks and descriptors}
\label{sec:repbaseline}
By using the same tracks and features as input, the previous sections have shown our RecLNet is able to correct the temporal localization weakness of state-of-the-art methods.
Here, we substitute these tracks and features (introduced in Section~\ref{sec:tubenets} and~\ref{sec:datasets}) by improved models. In the following, we first explain the track modification then the features substitution and finally analyze their impact.

\smallskip

\noindent{\bf Person action tracks with temporal integration.}
In order to improve person tracking, we integrate temporal information in the detector.
Supported by our analysis in Section~\ref{sec:explain_gain_temp}, showing that better spatial localization would greatly improve our results, and by~\cite{kalogeiton17iccv} that shows large detection improvement by stacking features coming from several neighboring frames, we design a new approach for tracks. In the same spirit as~\cite{kalogeiton17iccv} who stacks frame features in the SSD detector~\cite{liu2016ssd}, we adapt Faster-RCNN~\cite{ren2015faster} to perform accurate detection by temporally integrating stacks of $K$ images.
We introduce the following modifications to the Faster-RCNN pipeline. First, the inference pass, producing the feature map on which the ROI-pooling applies, is performed independently on $K$ consecutive frames. The $K$ output feature maps are stacked along the channel dimension and will serve at computing scores and regressions. Second, the RPN computes at each anchor location a global objectness score for the stack and $K$ regressions. Third, anchors labels are computed based on the mean overlap between the $K$ ground-truth boxes and the $K$ regressed proposals. Fourth, the 3D proposals are further regressed to action proposals (tubelets) by computing one action score and $K$ regressions per class.
Tubelets are linked into action tracks using the code of~\cite{kalogeiton17iccv}. This is an online method that iteratively aggregates tubelets sorted by action score to a set of current links based on spatio-temporal overlap.
The detector is used on the ResNet-101 architecture~\cite{he2016deep} and we choose $K=5$ since it is a good trade-off between tubelet quality and efficiency. Since only sparse keyframes are spatially annotated on DALY, automatic tracking is performed to propagate the ground truth (see Section~\ref{sec:datasets_daly}). Here, stacks used at train time on DALY are then composed of one annotated keyframe and its 2 frames tracked forward and backward. These action tracks are referred to as \emph{stack} in the following.

\noindent{\bf I3D features.}
Recent results have shown large action recognition improvement using the I3D~\cite{carreira2017quo} architecture.
We replace the per-frame features by descriptors extracted with the I3D RGB and flow networks both trained on the Kinetics dataset~\cite{kay2017kinetics}.
We extract features after the 7-th inception block, before the max-pooling, as it is a good balance between deepness, for strong classification results, and accurate resolution, for precise track pooling. After this block, the temporal receptive field is roughly one hundred frames.
As input, we use a spatial resolution of $320 \times 240$ pixels resulting in feature maps of size $20 \times 15$ with $832$ channels.
These feature maps are extracted at intervals of 4 frames. The box temporally aligned with the middle of the interval is used as pooling region
to obtain a $832$-dimensional I3D descriptor representing the track at this time step.

\begin{table}[H]
\setlength{\tabcolsep}{1.2mm}
\def\arraystretch{1.2}
    \begin{center}
        \begin{tabular}{lcccccccccc}
            & &&&  \multicolumn{3}{c}{UCF101-24} &&  \multicolumn{3}{c}{DALY}\\       
            method & feat. & tracks & & 0.3 & 0.4 & 0.5 && 0.1 & 0.2 & 0.3
          	\\
            \hline            \hline
             RecLNet  	& {\scriptsize f. based}	& {\scriptsize f. based}	&& 69.0 & 57.5 & 46.5  && 30.2 & 25.4 & 19.7
            \\ RecLNet 	& I3D					& {\scriptsize f. based}	&& 72.3 & 63.2 & 50.8  && 37.9 & 33.9 & 26.8
            \\ RecLNet 	& I3D       				& \emph{stack}			&& \textbf{77.4} & \textbf{68.5} & \textbf{57.5}
            																&& \textbf{41.1} & \textbf{37.6} & \textbf{31.0}
            \\
            \hline
            FC 	        & I3D 					& {\scriptsize f. based}&&  71.8 & 62.3 & 49.8 && 35.5 & 32.1 & 25.0
             \\FC       & I3D 					&  \emph{stack}			&& 75.4 & 67.4 & 56.3 && 38.7 & 34.9 & 28.6
        \end{tabular}
    \end{center}
    \caption{Spatio-temporal action localization (mAP) performance on UCF101-24 (at IoU 0.3, 0.4 and 0.5) and
			DALY (at IoU 0.1, 0.2 and 0.3) when substituting the frame-based features by I3D and the frame-based tracks by \emph{stack}.
			Results are reported for RecLNet and the fully connected classifier (FC).
     \label{tbl:impbaseline}}
\end{table}
\noindent{\bf Experimental evaluation.} Here, frame-based features and frame-based tracks we refer to are the ones described in Section~\ref{sec:datasets}.
Table~\ref{tbl:impbaseline} first shows that when substituting the baseline features by the I3D descriptors our RecLNet model obtains a performance boost of around $4\%$ on UCF101-24 ($+4.3\%$ at IoU 0.5) and $7$-$8\%$ on DALY ($+7.1\%$ at IoU 0.3). This result is inline with~\cite{gu2017ava}. When we further replace the frame-based tracks by ours obtained with stack-Faster-RCNN (\emph{stack}), RecLNet results get a second improvement of $5$-$7\%$ on UCF101-24 and $3$-$4\%$ on DALY. This gain of performance validates what was shown by~\cite{kalogeiton17iccv}, i.e, integrating temporal information in the detector increases the track spatial precision leading to action localization improvement (as we observed in Section~\ref{sec:explain_gain_temp}).
For comparison we also report the fully connected classifier (FC) results as in Section~\ref{sec:exp_imp}. Given the large temporal receptive field of I3D features mentioned earlier, we notice RecLNet still improves the performance, e.g, it gets $+2.0\%$ and $+2.4\%$ mAP at IoU 0.3 when using the \emph{stack} tracks  on UCF101-24 and DALY respectively. In the following, the RecLNet model with the I3D features and the improved tracks (\emph{stack}) is referred to as RecLNet++.

\subsection{Comparison to the state of the art}
\label{sec:comp}

\begin{table}
    \setlength{\tabcolsep}{1.3mm}
    \def\arraystretch{1.1}
    \centering 
    \begin{tabular}{lccccccccc}
       IoU && 0.05 & 0.1 & 0.2 & 0.3 & 0.4 & 0.5 & 0.6 & 0.75\\
        \hline        \hline
        \cite{Weinzaepfel_2015_ICCV} && 54.3 & 51.7 & 46.8 & 37.8 & - & - & - & -\\ 
        \cite{hou2017tube} && 54.7 & 51.3 & 47.1 & 39.2 & - & - & - & -\\
        \cite{zolfaghari2017chained} && 65.2 & 59.5 & 47.6 & 38.0 & - & - & - & -\\
        \cite{peng2016multi} && 54.5 & 50.4 & 42.3 & 32.7 & - & - & - & -\\
        \cite{weinzaepfel2016towards} && 71.1 & - & 58.9 & - & - & - & - & -\\
        \cite{peng2016hal} && 78.8 & 77.3 & 72.9 & 65.7 & - & - & - & -\\
        \cite{saha2016deep} &&  79.1 & 76.6 &  66.8 & 55.5 &  46.4 &  35.9 & 26.8 & -\\
        \cite{mettes2016} && - & - & 34.8 & - & - & - & - & -\\
        \cite{singh2017}  && - & - & 73.5 & - & - & 46.3 & - & 15.0\\
        \cite{saha2017amtnet} && - & 71.3 & 63.1 & 51.6 & - & 33.1 & - & -\\
        \cite{kalogeiton17iccv} && - & - & 76.5 & 65.2 & - & 49.2 & - & 19.7\\
        	\cite{gu2017ava} && - & - & - & - & - & \textbf{59.9} & - & -\\
        \hline
        RecLNet                  && 83.0 & 81.7 & 77.0 & 69.0 & 57.5 & 46.5 &  36.7 & 10.3 \\
        RecLNet++                && \textbf{86.6} & \textbf{86.1} & \textbf{83.4} & \textbf{77.4} & \textbf{68.5} & 57.5 & \textbf{46.0} & \textbf{23.9} \\
  
    \end{tabular}
    \vspace{2mm}
    \caption{
        Comparison to the state of the art on UCF101-24 (mAP for IoU
        values ranging from 0.05 to 0.6). 
        \label{tbl:sota}}
\end{table}

\smallskip
\noindent{\bf UCF101-24.}
Table~\ref{tbl:sota} compares RecLNet and RecLNet++ models to~\cite{gu2017ava,hou2017tube,kalogeiton17iccv,mettes2016,peng2016multi,peng2016hal,saha2016deep,saha2017amtnet,singh2017,Weinzaepfel_2015_ICCV,weinzaepfel2016towards,zolfaghari2017chained}
on UCF101-24 spatio-temporal action localization. 
RecLNet is our recurrent network with GRU memory units and \emph{fusion layer} to combine optical flow and appearance features performing localization with our \emph{threshold} approach
and RecLNet++ uses I3D features and our improved \emph{stack} tracks.
RecLNet already significantly outperforms all other methods for all
IoU thresholds below $0.5$.
Due to our significantly improved temporal localization,
our approach outperforms~\cite{saha2016deep} on which we build
by $+9.9\%$ at high IoU ($0.6$).
We also outperform~\cite{peng2016hal}, despite the fact that
they use more sophisticated features combining a number of different
human parts as well as multi-scale training and testing. Both these
components are complementary to our approach.
Our significant boost in performance can be explained by the fact
that most current action localization methods rely on spatial
features, for example per-frame CNN descriptors for optical
flow and appearance, to detect action both spatially and
temporally. However, such features are not designed for temporal
localization. By relying on an accurate temporal model such as our
recurrent RecLNet, we can clearly improve temporal detection and
methods we build on that do not focus on this aspect.
Recent method~\cite{kalogeiton17iccv}, relying on more accurate
spatial localization obtained by spatio-temporal cuboid regression,
outperforms our model at high IoU. Indeed, they benefit from
more precise human track bounding-boxes than the one we build on.
Such an approach is used by RecLNet++, the enhanced version of RecLNet. This is shown to improve the state of the art by a large margin outperforming~\cite{kalogeiton17iccv} by $+12.2\%$, $+8.3\%$ and $+4.2\%$ at IoU 0.3, 0.5 and 0.75 respectively.
We also mention very recent work~\cite{gu2017ava} which only reports results at IoU 0.5 and achieves similar performance.

\begin{table}
\setlength{\tabcolsep}{2mm}
\def\arraystretch{1.2}
    \begin{center}
        \begin{tabular}{lcccccc}
            {scores} && {localization}  && 0.1 & 0.2 & 0.3 \\
            \hline
            \hline
            \cite{weinzaepfel2016towards} && \emph{sliding window}  && - & 14.5  & - \\
            CNN RGB+OF &&  \emph{threshold}   && 22.9 & 17.8  & 13.4 \\
            \hline
            RecLNet  &&  \emph{threshold}  && 30.2  &  25.4  &  19.7 \\
            RecLNet++  &&  \emph{threshold}  && \textbf{41.1}  &  \textbf{37.6}  &  \textbf{31.0} \\
            \hline
                        
        \end{tabular}
    \end{center}
    \caption{State of the art on DALY (mAP for IoU 0.1, 0.2 and 0.3). Localization  indicates the method used for temporal localization.   \label{tbl:daly}}
     
\vspace{-3mm}
\end{table}

\begin{table}
\newcommand\sihnum{\small}

\def\arraystretch{1.2}
    \begin{center}
        \begin{tabular}{lcccc}
         	Action classes&& CNN RGB+OF && RecLNet\\
            \hline
            \hline
           \emph{App. MakeUp Lips}               &&\sihnum 3.4&&\sihnum \textbf{13.6}	\\
           \emph{Brushing Teeth}           	 &&\sihnum 7.9&&\sihnum \textbf{19.3}	\\
           \emph{Cleaning Floor}      &&\sihnum 14.3&&\sihnum \textbf{24.7}	\\
           \emph{Cleaning Windows}    &&\sihnum 6.5&&\sihnum \textbf{10.9}	\\
           \emph{Drinking}            &&\sihnum \textbf{20.8}&&\sihnum 13.1	 \\
           \emph{Folding Textile}     &&\sihnum 7.2&&\sihnum \textbf{16.6}	\\
           \emph{Ironing}             &&\sihnum 25.4&&\sihnum \textbf{32.7} 	\\
           \emph{Phoning}             &&\sihnum 7.6&&\sihnum \textbf{19.9}	 \\
           \emph{Playing Harmonica}   &&\sihnum 33.4&&\sihnum \textbf{34.4}	\\
           \emph{Taking Photos/Videos}&&\sihnum 7.3&&\sihnum \textbf{12.2}	\\
          \hline
          mAP               			 &&  13.4 && \textbf{19.7} \\  

        \end{tabular}
    \end{center}
    \caption{Per-class performance on DALY (mAP at IoU 0.3).
        \label{tbl:DALYperClass}}
\end{table}

\smallskip
\noindent{\bf DALY.}
The recent DALY dataset~\cite{weinzaepfel2016towards} is very challenging for spatio-temporal localization because
it contains only short actions compared to the video length making temporal segmentation crucial and difficult.
Table~\ref{tbl:daly} compares our results to the state of the art.
We can see that our RecLNet approach significantly outperformst~\cite{weinzaepfel2016towards} by more than $10\%$. 
We also report results with our CNN scores~\emph{CNN RGB+OF} and temporal thresholding of the action scores
without using RecLNet. Interestingly, the baseline approach outperforms the state of the art, but performs
significantly worse than our (\emph{RecLNet}) method. Also, Table~\ref{tbl:DALYperClass} shows RecLNet improves over the baseline for all
classes except \emph{Drinking}.
Again, since we build on tracks from~\cite{weinzaepfel2016towards} and extract
similar features (see Section~\ref{sec:method}), the improvement should be attributed to the more accurate temporal localization
achieved by our RecLNet model. RecLNet++ further improves the results and outperforms~\cite{weinzaepfel2016towards} by $+23.1\%$.

\vspace{-2mm}
\subsection{Qualitative results}
\label{sec:qualitative}
Good temporal localization needs precise evolution of action score with clear temporal boundaries.
Having smooth scores also help our \emph{threshold} localization method presented in Section~\ref{sec:postproc}.
Here, we provide qualitative results to illustrate that our RecLNet possesses these properties
and is well-suited for such localization.

Figure~\ref{scoreblobs} shows scores from~\cite{saha2016deep} (red
curves) versus our RecLNet response (green curves) for the 5 action track
proposals used on UCF101-24. Thus, there are in total 10 curves per
graph, one curve per track for the two compared methods. A bold curve
(resp.\ dashed curve) shows that its corresponding track overlaps with
more (resp. less) than $0.5$ spatial IoU with a ground-truth action
annotation. The ground-truth action interval is represented by the
horizontal blue bar below the x-axis. The x-axis represents time. For each graph, we show the
frame corresponding to the maximum RecLNet response (localized by  the
vertical yellow line) with its associated human detection (yellow
box). 

We show  scores for 3 short actions \emph{basketball}, \emph{tennis
  swing} and \emph{volleyball spiking}, and 2
long actions, \emph{cliff diving} and \emph{diving}. 
We can observe that the scores of the RecLNet correspond to the temporal
boundaries more precisely than the CNN scores of~\cite{saha2016deep}. 
Scores from~\cite{saha2016deep} are often flat and have
high values throughout the entire video length (Figure~\ref{scoreblobs},
row 3). In general, they are not precise temporally (e.g.,
Figure~\ref{scoreblobs}, row 4). Also, a lot of dashed lines are highly
scored (Figure~\ref{scoreblobs}, rows 3 and 5) which adds false positives. 
Finally, these graphs show that it is easy to set a threshold on RecLNet
responses in order to obtain good temporal localization, while scores
from~\cite{saha2016deep} (and scores from CNNs in general) are often
imprecise temporally.

Similarly, Figures~\ref{scoreblobs1}, \ref{scoreblobs2} show qualitative results for temporal action localization on the DALY dataset, comparing again our RecLNet (green curves) and the \emph{CNN RGB+OF} baseline (red
curves).
Each curve corresponds to one person track in the video.
Each row corresponds to one action class.
The last (third) column in Figures~\ref{scoreblobs1}, \ref{scoreblobs2} corresponds to failure cases or examples where our RecLNet model does not improve temporal localization.
Similarly to results for the UCF101-24 dataset, the two first columns shows that scores of RecLNet are better aligned with the ground truth action boundaries compared to scores of the \emph{CNN RGB+OF} baseline.
At the same time, dashed lines indicate many high-scoring false positive detections for the \emph{CNN RGB+OF} method (e.g.~Figure~\ref{scoreblobs1}, row 1, plot 2 and~Figure~\ref{scoreblobs2}, row 2, plot 2). The last column in Figures~\ref{scoreblobs1}, \ref{scoreblobs2} indicates that DALY is a challenging dataset with many difficult examples where, most of the time, both methods fail to detect the correct temporal action boundaries.
 
\vspace{-2mm}
\section{Conclusion}
\label{sec:conclusion}
This paper shows that training a recurrent model (GRU) on the level of person tracks for modeling the temporal structure
of actions improves localization in time significantly. Building on
current state-of-the-art methods that obtain very good results for 
spatial localization, RecLNet improves significantly the detection of 
action time boundaries and hence the overall performance of spatio-temporal
localization. As demonstrated in our analysis, improving spatial precision of human
tracks, inspired by most recent methods, and using better features further improve our
results (RecLNet++).
Our method outperforms the state of the art on the two
challenging datasets, namely UCF101-24 and DALY.

\section*{Acknowledgements}
This work was supported in part by ERC grants ACTIVIA and ALLEGRO, the MSR-Inria joint lab, the Louis Vuitton ENS Chair on Artificial Intelligence, an Amazon academic research award, and an Intel gift.

\newcommand\figy{34mm}

\begin{figure*}
    \begin{tabular}{ccc}

        \includegraphics[width=55mm,height=\figy]{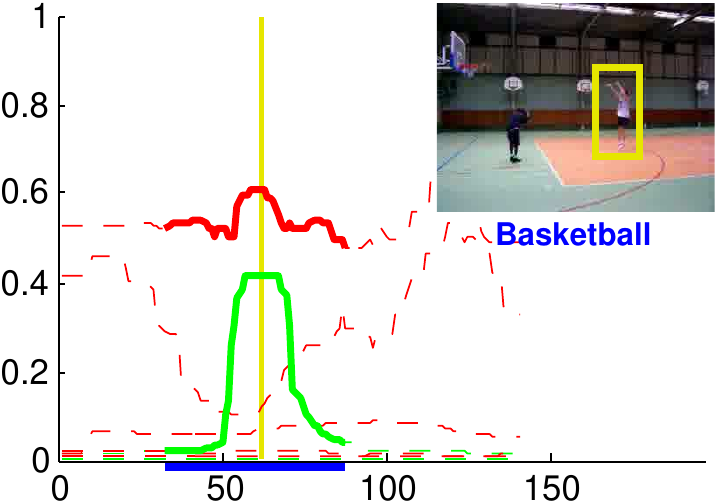}
         & \includegraphics[width=55mm,height=\figy]{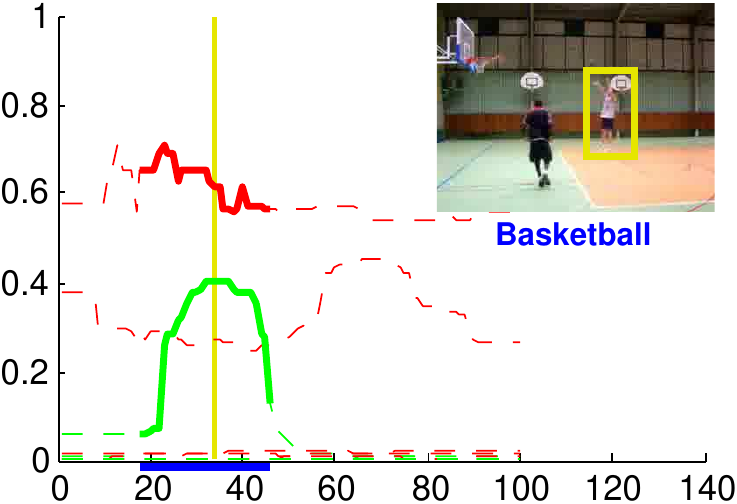}
         &  \includegraphics[width=55mm,height=\figy]{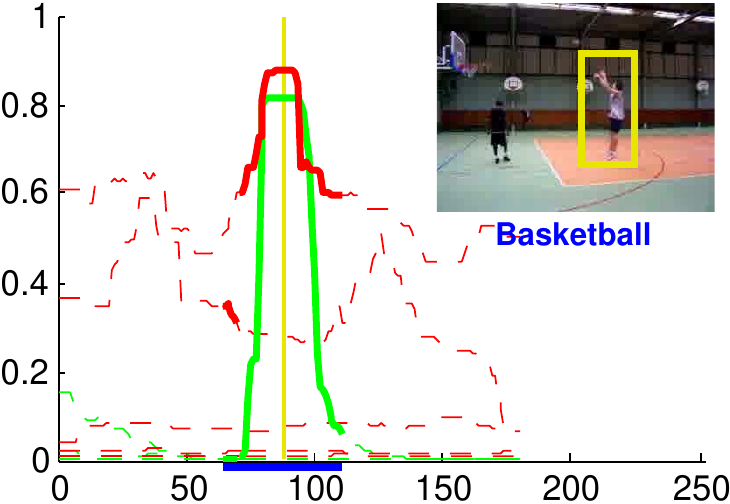}
         \\ \\
         \includegraphics[width=55mm,height=\figy]{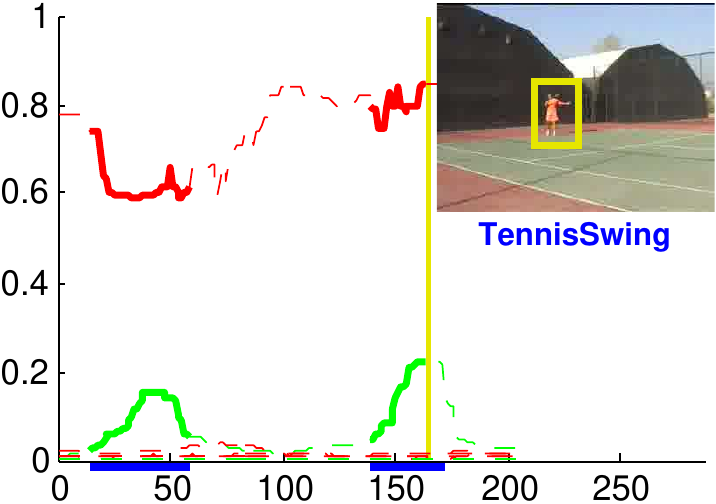}
        &\includegraphics[width=55mm,height=\figy]{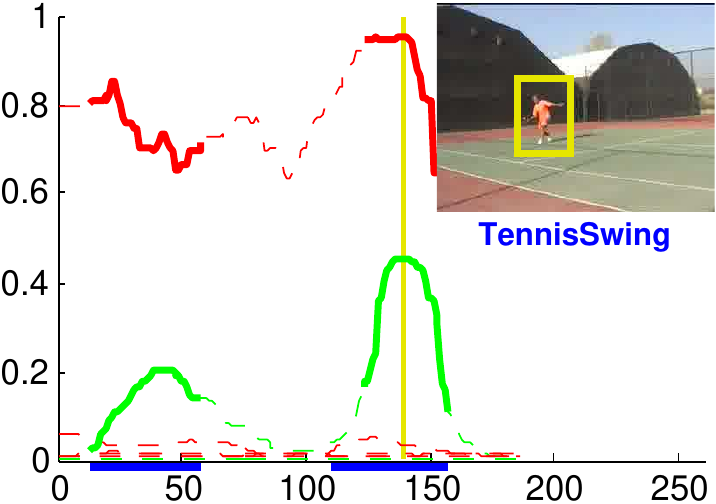}
         &  \includegraphics[width=55mm,height=\figy]{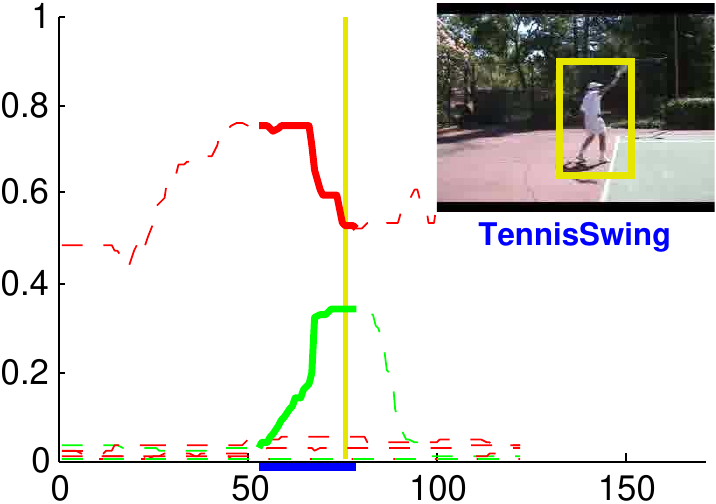}
         \\ \\

          \includegraphics[width=55mm,height=\figy]{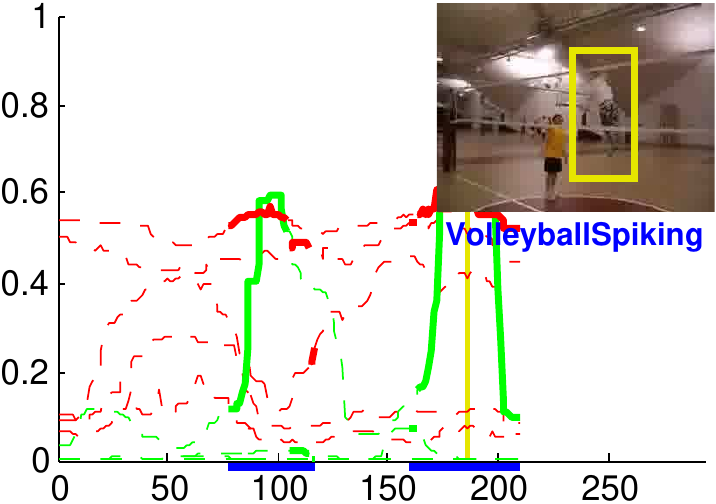}
        &\includegraphics[width=55mm,height=\figy]{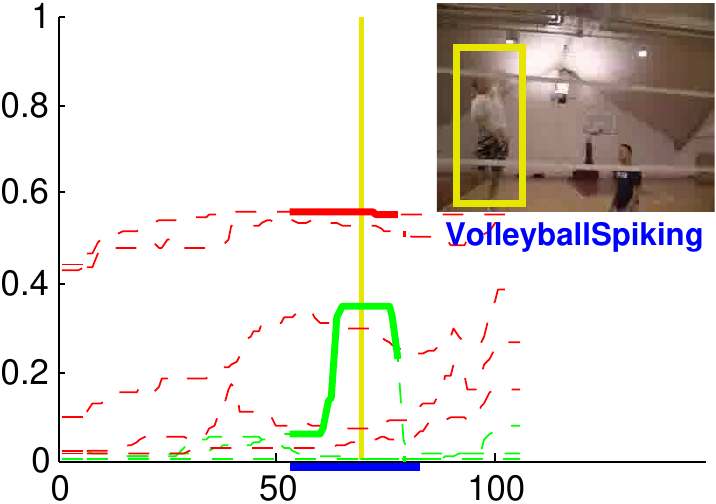}
         &  \includegraphics[width=55mm,height=\figy]{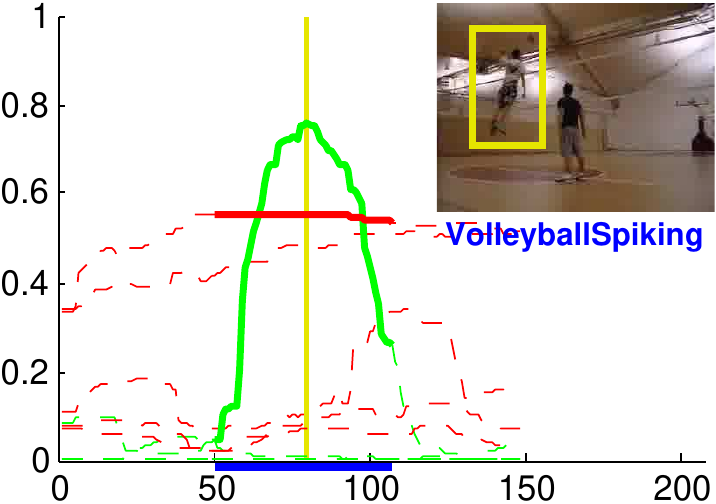}         
         \\ \\            
         
          \includegraphics[width=55mm,height=\figy]{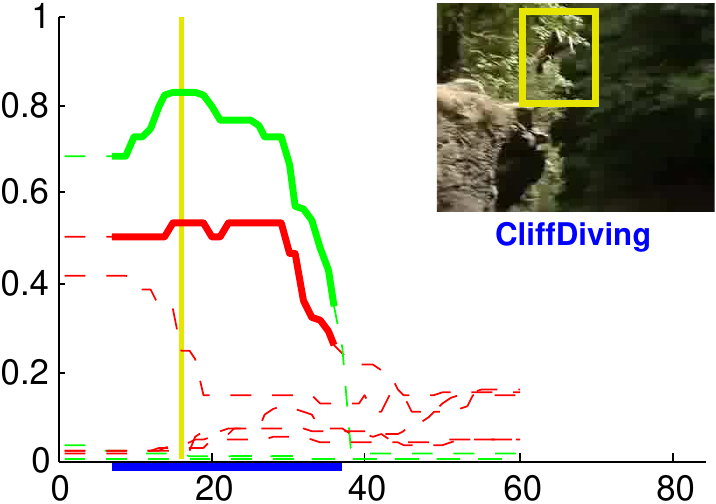}
         & \includegraphics[width=55mm,height=\figy]{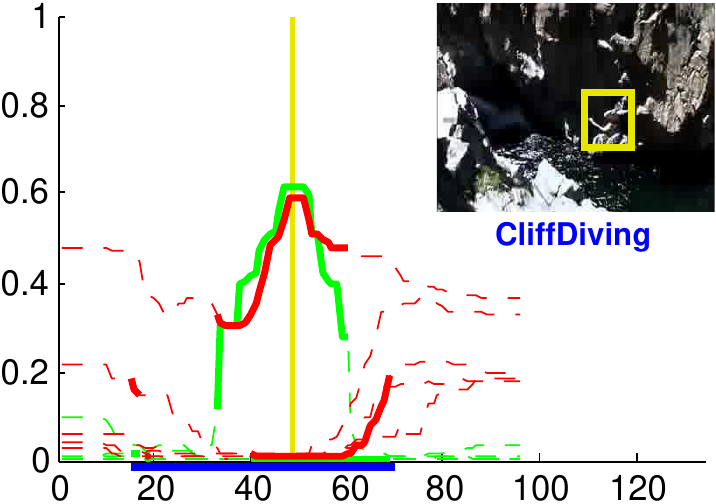}
         &  \includegraphics[width=55mm,height=\figy]{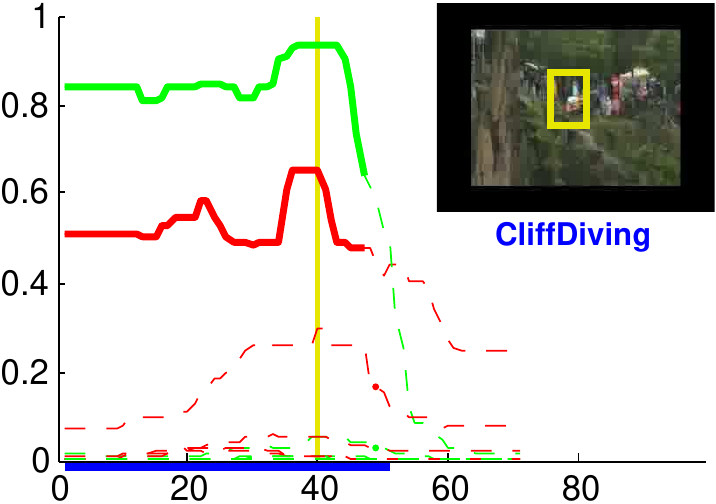}         
         \\ \\ 
          \includegraphics[width=55mm,height=\figy]{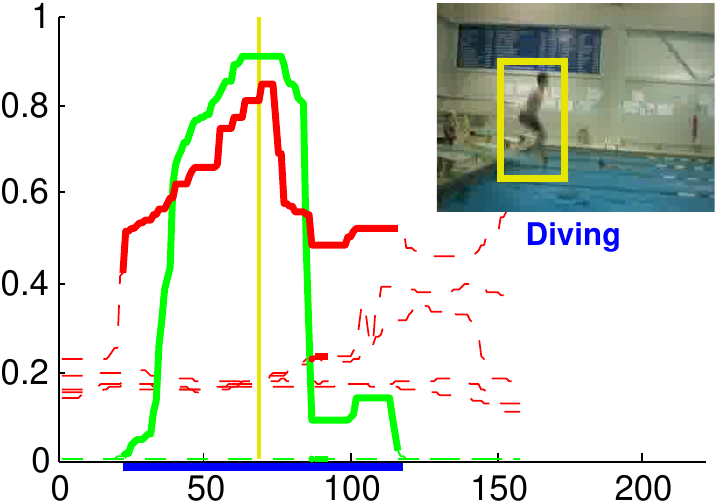}
        &\includegraphics[width=55mm,height=\figy]{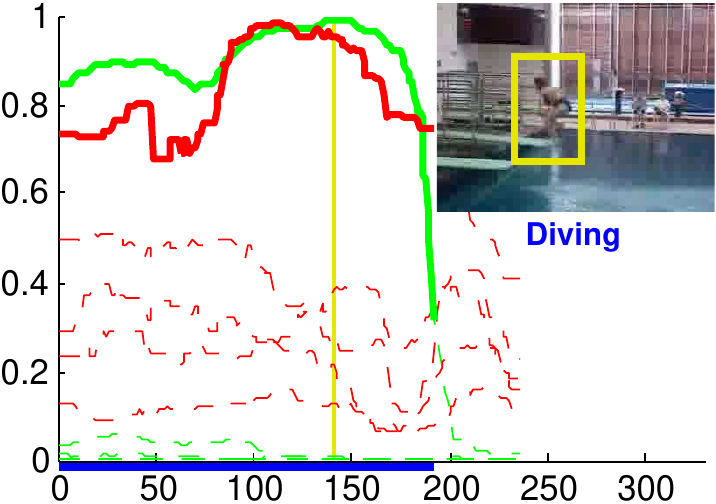}
         &  \includegraphics[width=55mm,height=\figy]{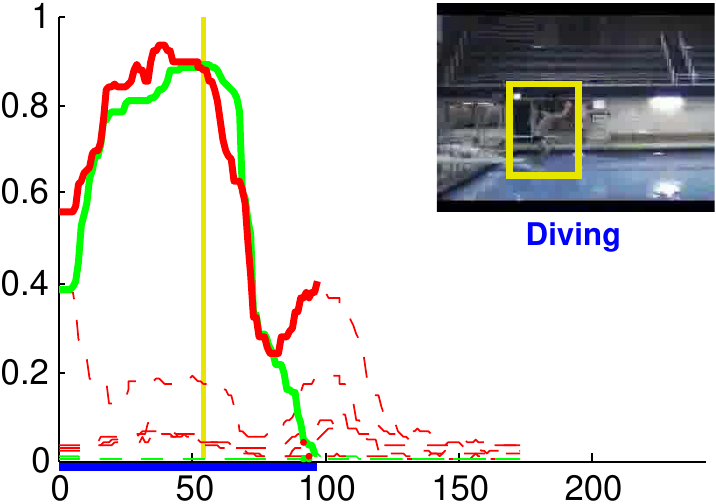}
    \end{tabular}
    \caption{Qualitative results for temporal localization on UCF101-24. Each curve
      represents a human track, green curves correspond to our RecLNet
      scores and red ones to scores from~\cite{saha2016deep}. A
      bold curve (resp. dashed curve) shows that the track overlaps
      with more (resp. less) than $0.5$ spatial IoU with a
      ground-truth action. The horizontal blue bar represents the
      ground-truth time segment. The video frame corresponds to the
      maximum RecLNet response (localized at the vertical yellow line)
      with its associated human detection (yellow box). The x-axis represents frame numbers.
    \label{scoreblobs}}
\end{figure*}

\setlength{\fboxsep}{1pt}

\begin{figure*}
    \begin{tabular}{ccc}
        \includegraphics[width=55mm,height=\figy]{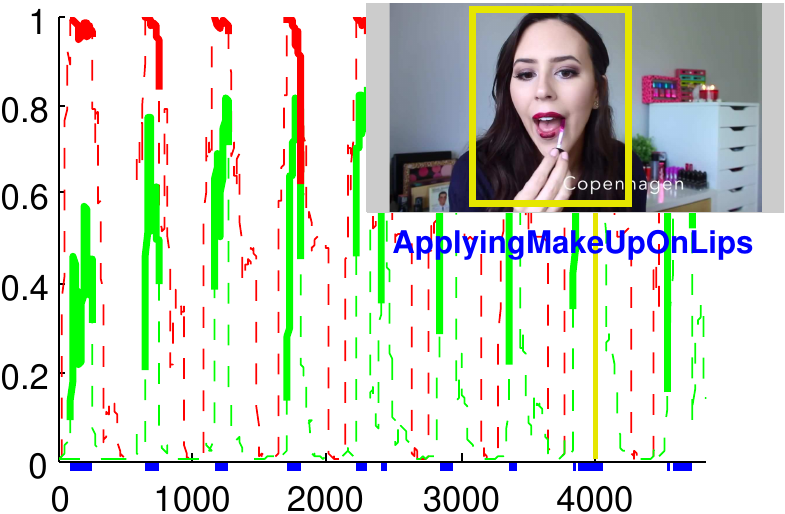}
         & \includegraphics[width=55mm,height=\figy]{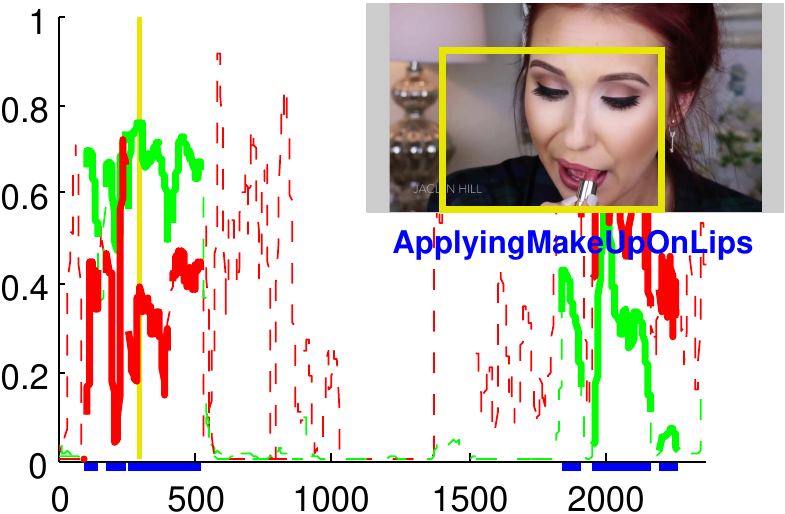}
         &  \color{red}{\fbox{\includegraphics[width=55mm,height=\figy]{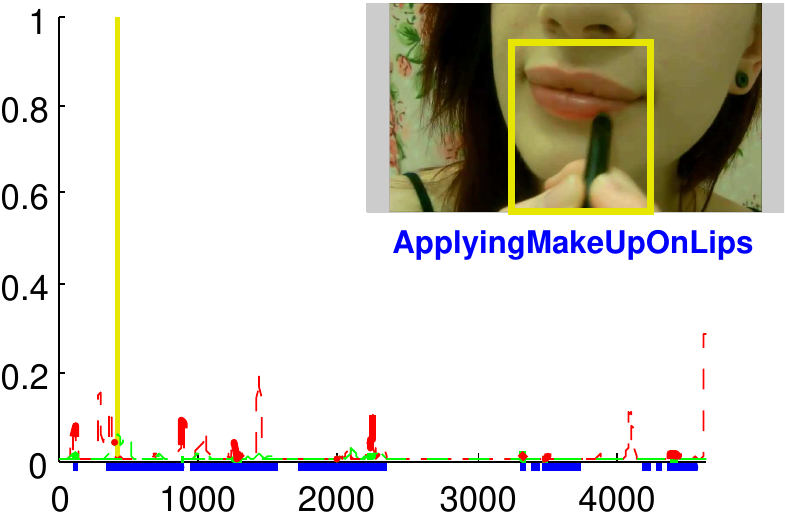}}}
         \\ \\
        \includegraphics[width=55mm,height=\figy]{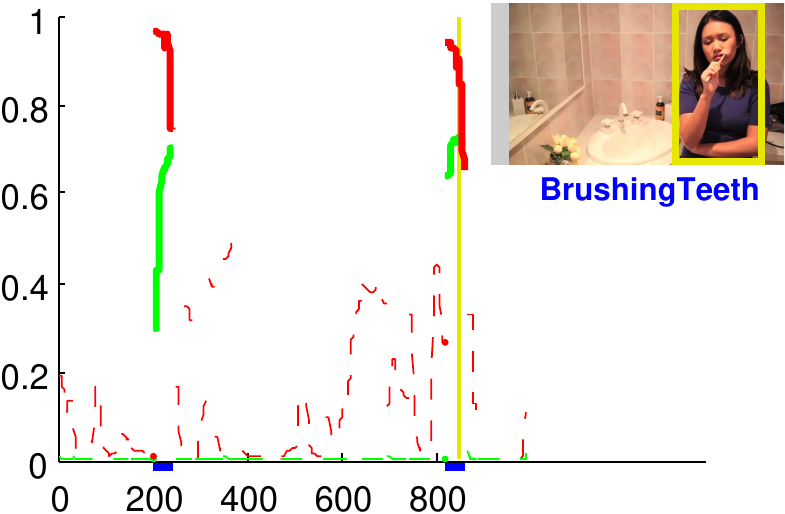}
         & \includegraphics[width=55mm,height=\figy]{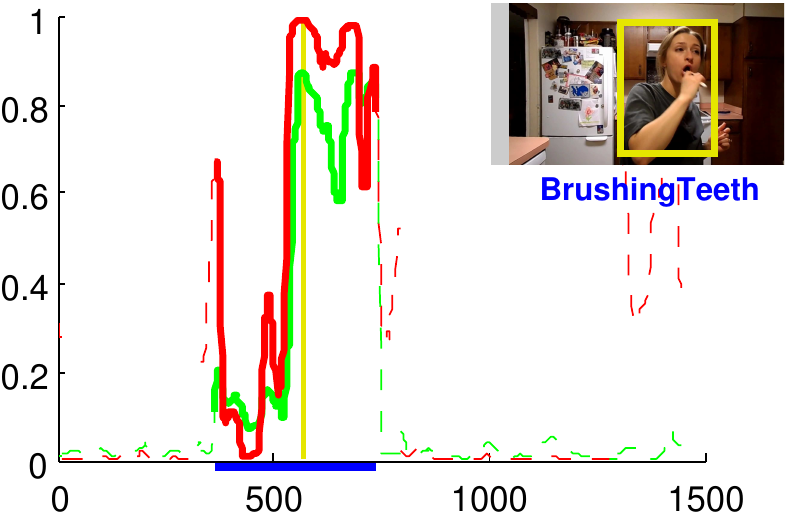}
         & \color{red}{\fbox{\includegraphics[width=55mm,height=\figy]{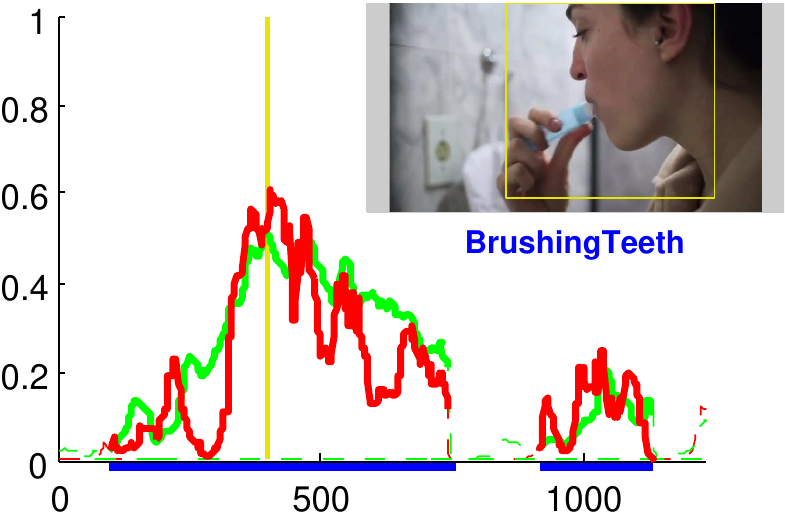}}}
         \\ \\     
        \includegraphics[width=55mm,height=\figy]{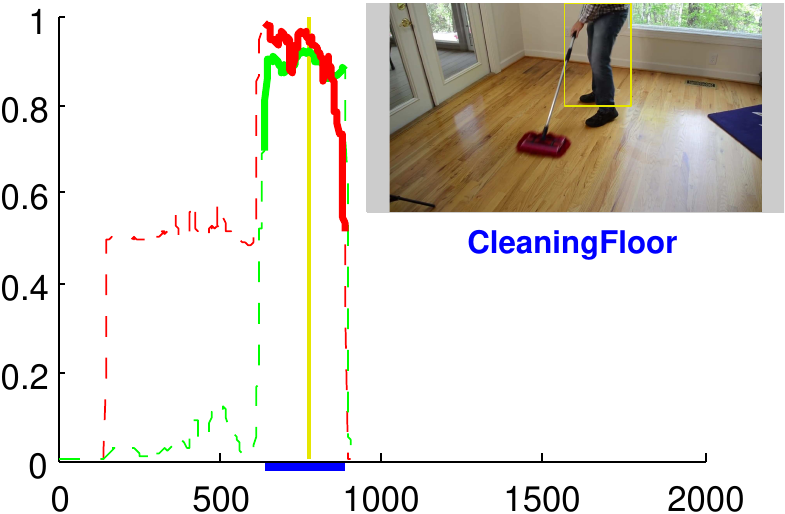}
         & \includegraphics[width=55mm,height=\figy]{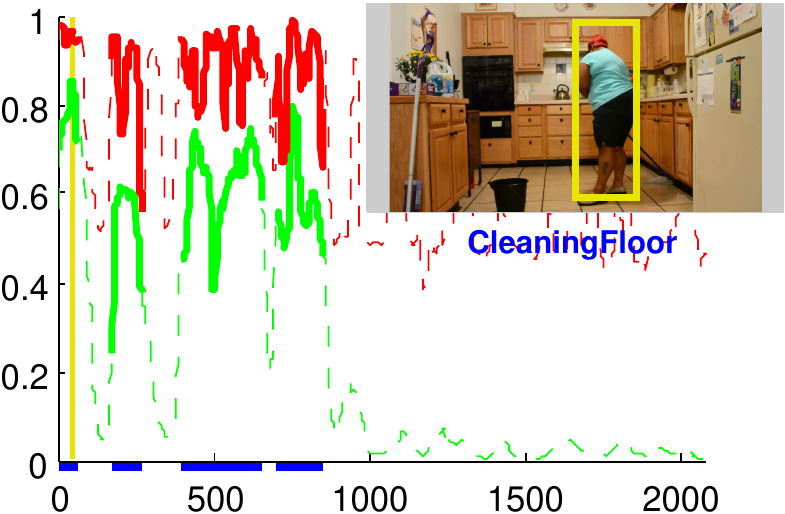}
         &  \color{red}{\fbox{\includegraphics[width=55mm,height=\figy]{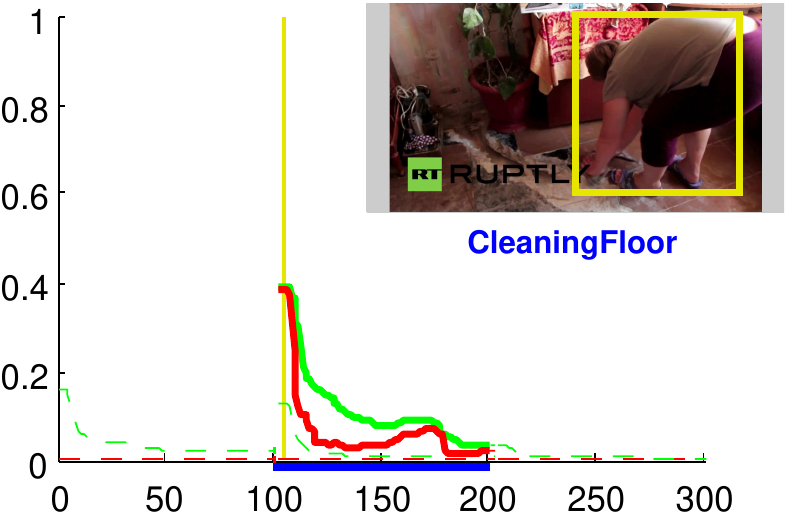}}}
         \\ \\
        \includegraphics[width=55mm,height=\figy]{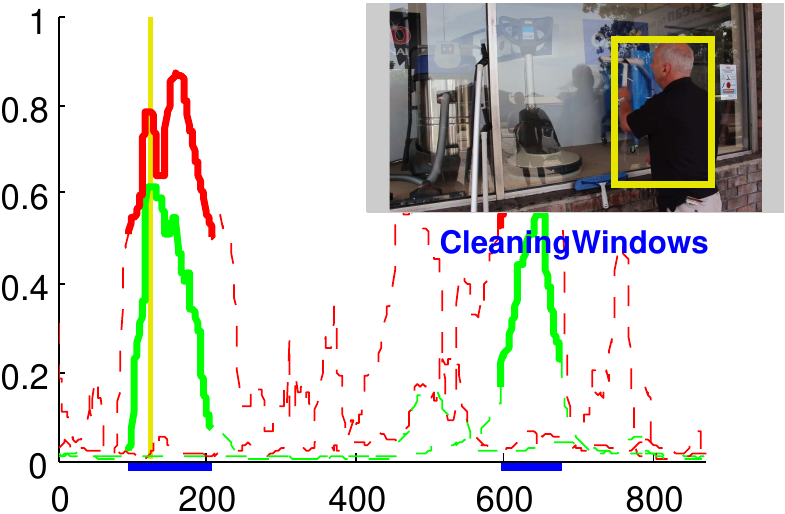}
         & \includegraphics[width=55mm,height=\figy]{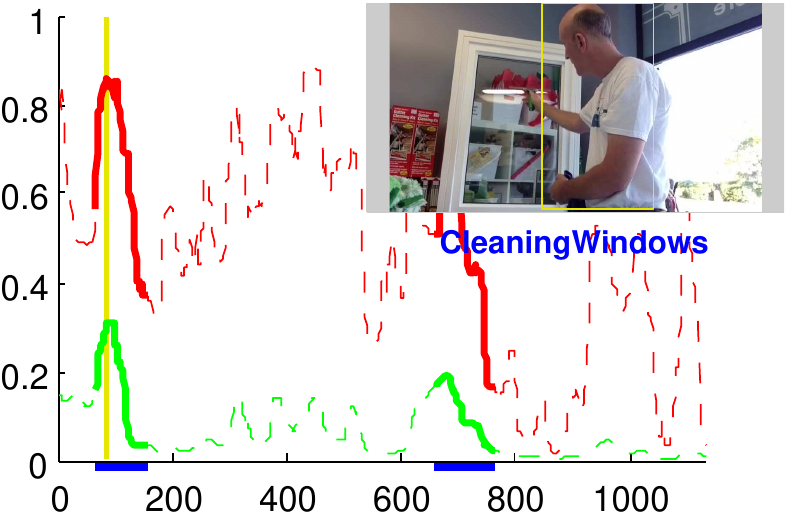}
         &  \color{red}{\fbox{\includegraphics[width=55mm,height=\figy]{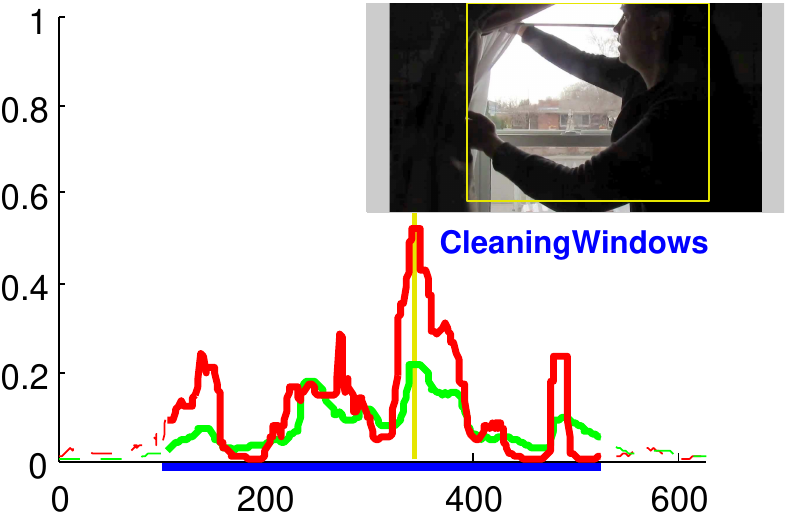}}}
         \\ \\        
        \includegraphics[width=55mm,height=\figy]{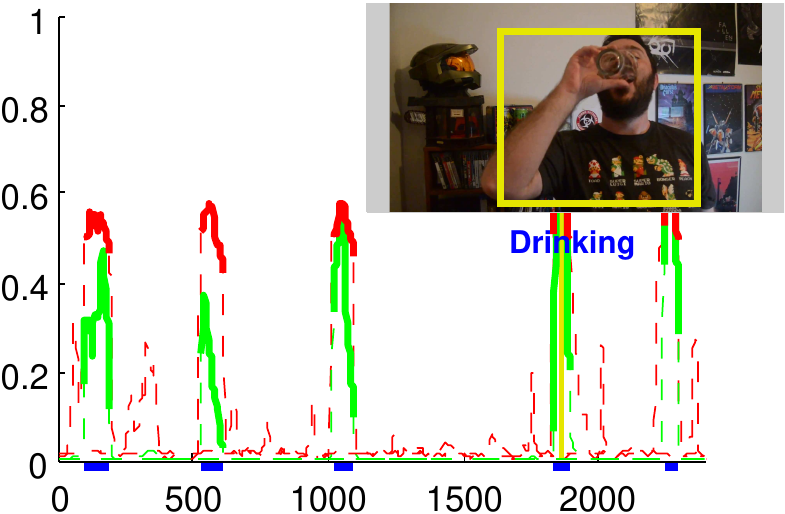}
         & \includegraphics[width=55mm,height=\figy]{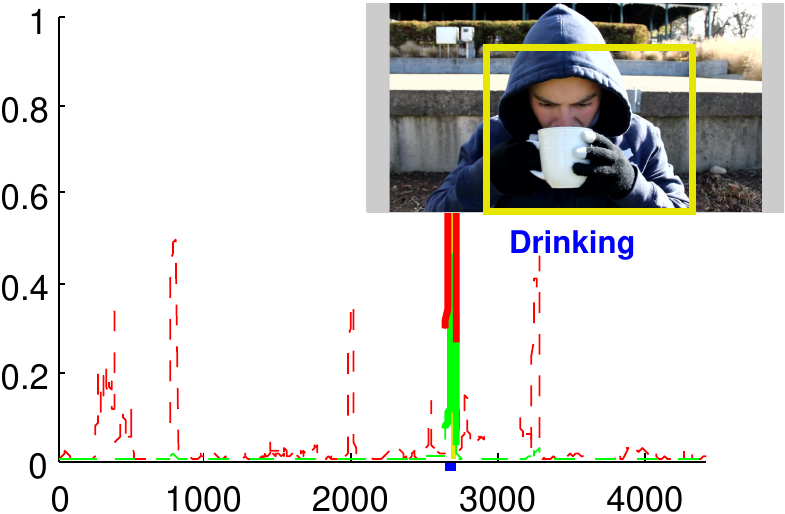}
         &  \color{red}{\fbox{\includegraphics[width=55mm,height=\figy]{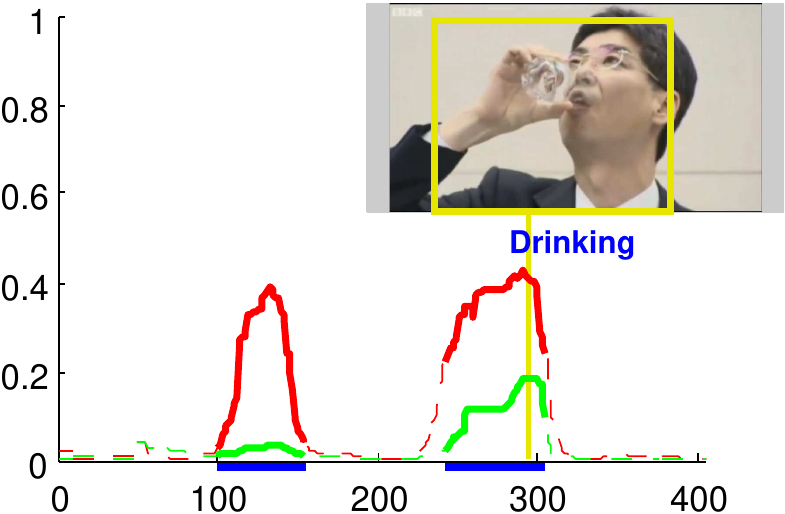}}}
    \end{tabular}
    \caption{Qualitative results for temporal localization on DALY. Each curve
      represents a human track, green curves correspond to our RecLNet
      scores and red ones to scores from CNN features~\emph{CNN RGB+OF}. A
      bold curve (resp. dashed curve) shows that the track overlaps
      with more (resp. less) than $0.5$ spatial IoU with a
      ground-truth action. The horizontal blue bar represents the
      ground-truth time segment. The video frame corresponds to the
      maximum RecLNet response (localized at the vertical yellow line)
      with its associated human detection (yellow box). The last column corresponds to failure cases or examples where our RecLNet model does not improve temporal localization. Each row corresponds to one action class, namely \emph{Applying Make Up On Lips}, \emph{Brushing Teeth}, \emph{Cleaning Floor}, \emph{Cleaning Windows} and \emph{Drinking}. The x-axis represents frame numbers.
    \label{scoreblobs1}}
\end{figure*}

\begin{figure*}
    \begin{tabular}{ccc}
        \includegraphics[width=55mm,height=\figy]{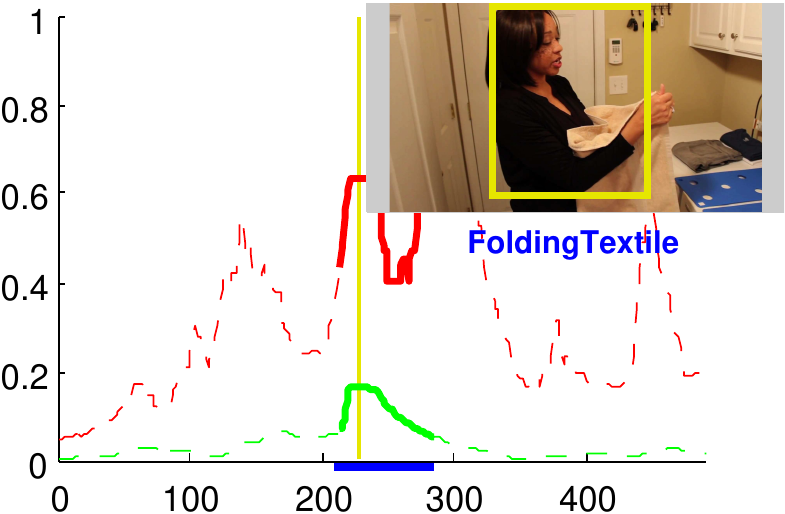}
         & \includegraphics[width=55mm,height=\figy]{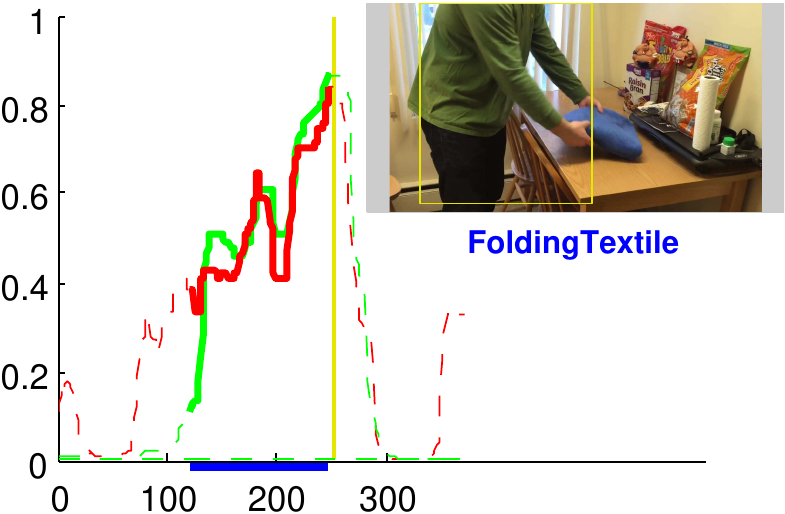}
         &  \color{red}{\fbox{\includegraphics[width=55mm,height=\figy]{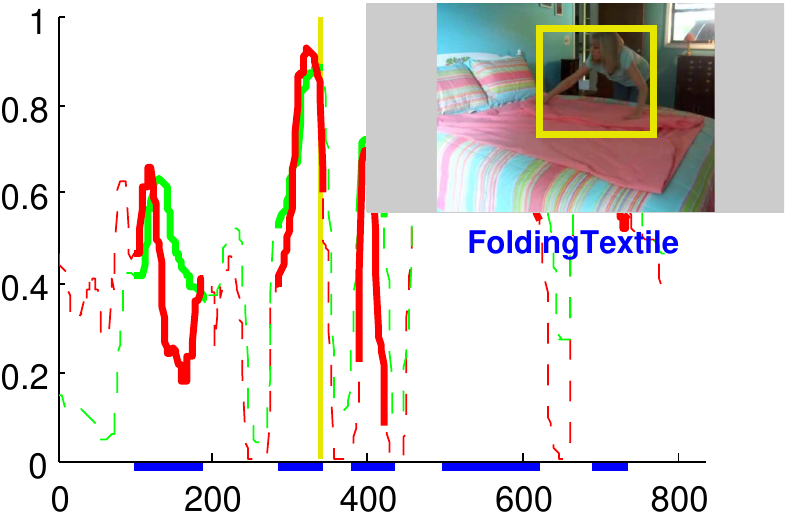}}}
         \\ \\
        \includegraphics[width=55mm,height=\figy]{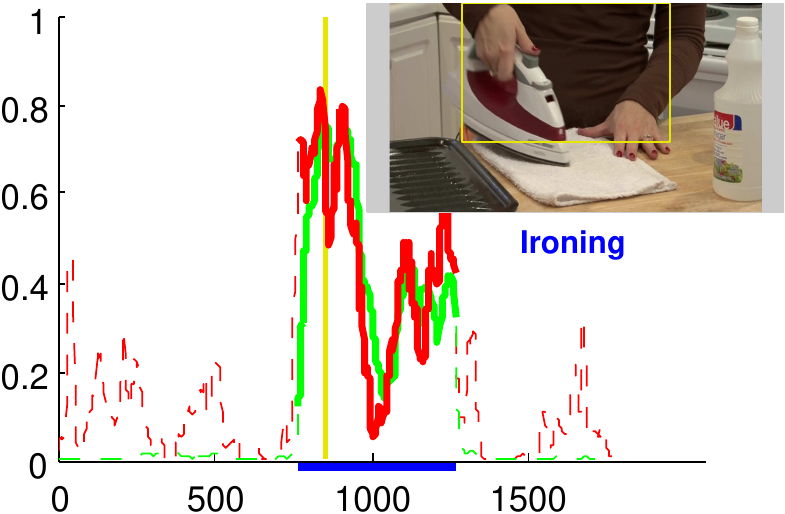}
         & \includegraphics[width=55mm,height=\figy]{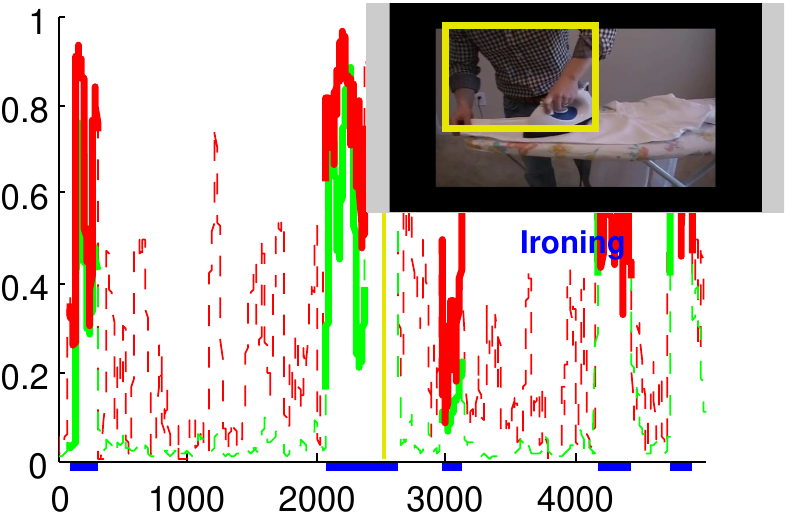}
         & \color{red}{\fbox{\includegraphics[width=55mm,height=\figy]{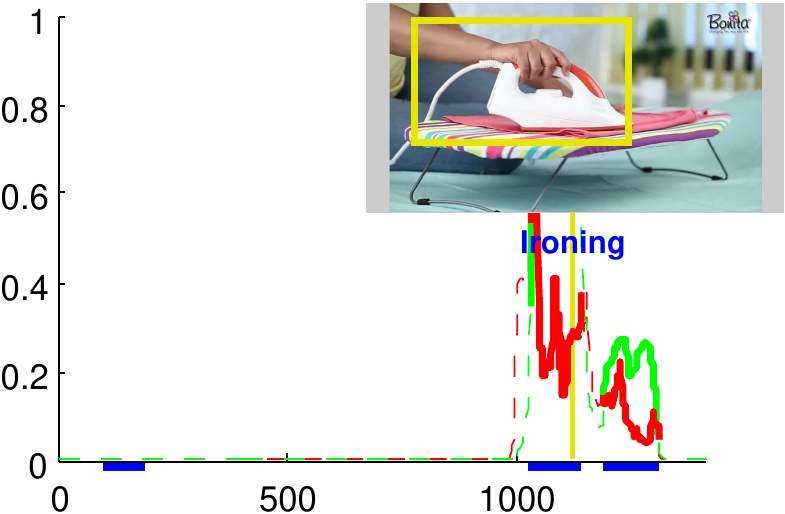}}}
         \\ \\     
         \includegraphics[width=55mm,height=\figy]{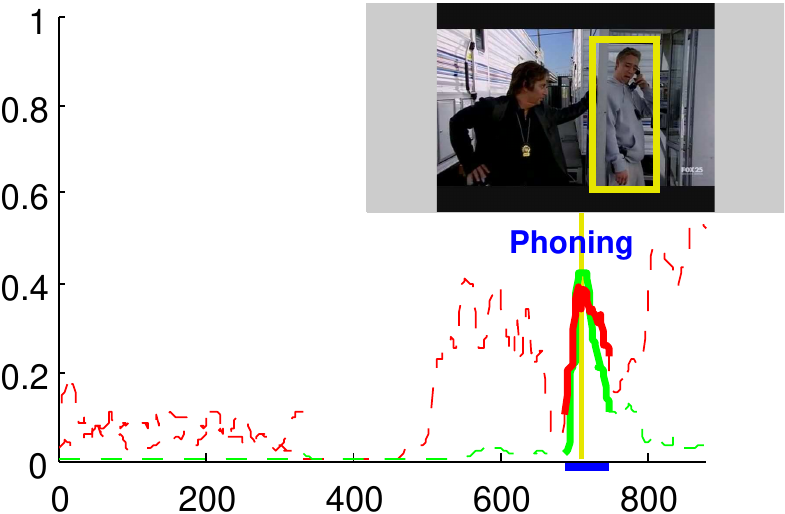}
         & \includegraphics[width=55mm,height=\figy]{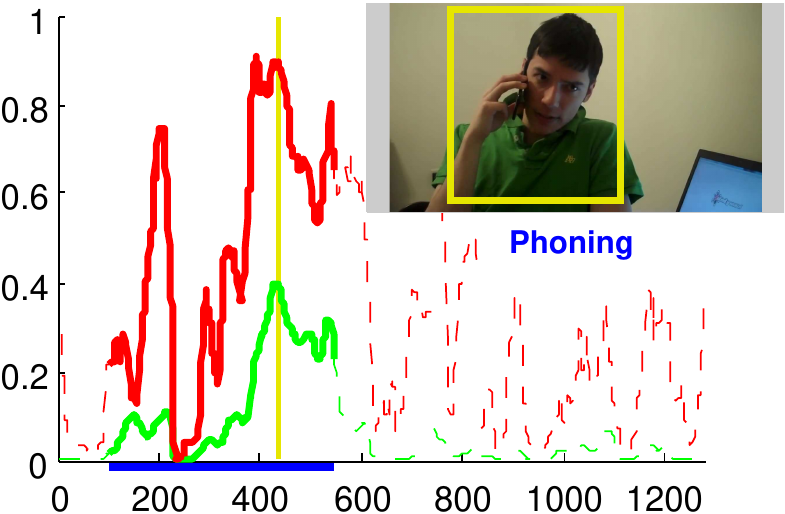}
         &  \color{red}{\fbox{\includegraphics[width=55mm,height=\figy]{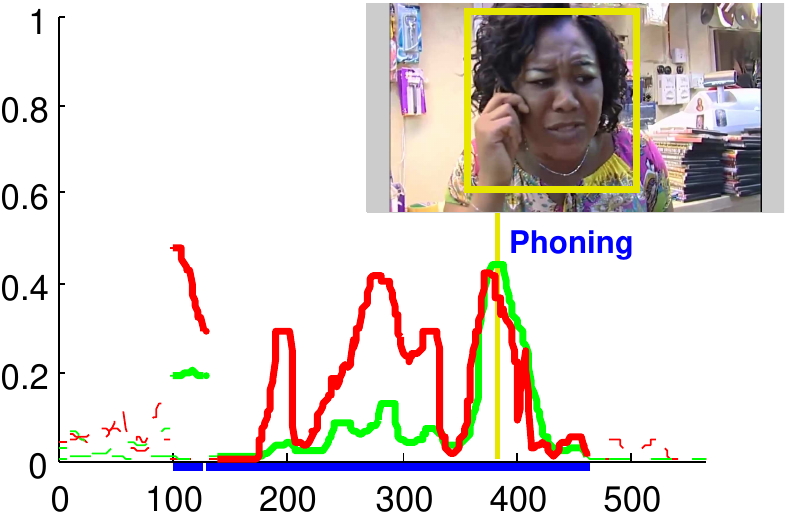}}}
         \\ \\
        \includegraphics[width=55mm,height=\figy]{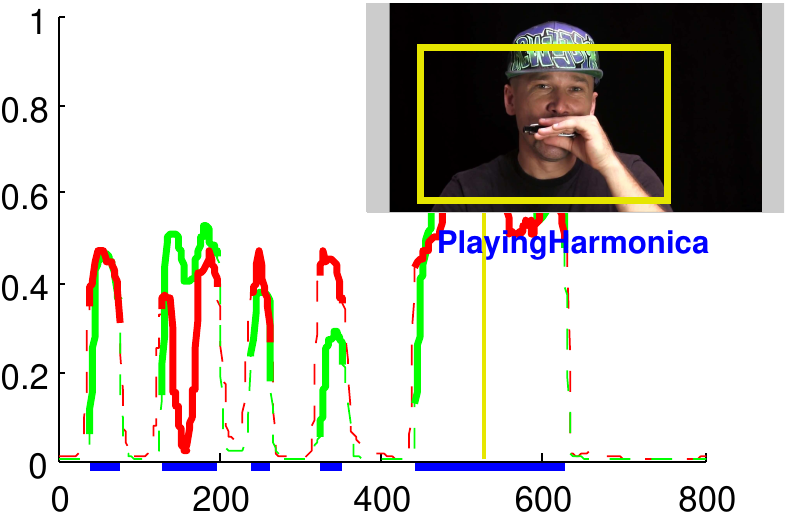}
         & \includegraphics[width=55mm,height=\figy]{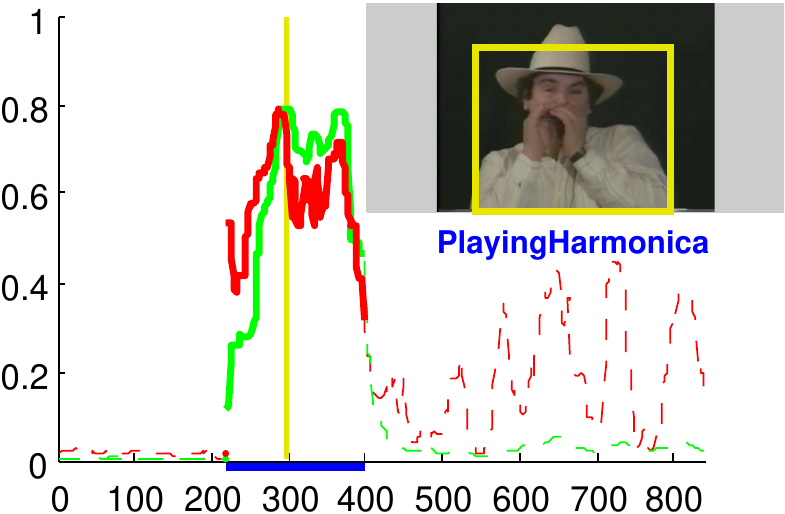}
         &  \color{red}{\fbox{\includegraphics[width=55mm,height=\figy]{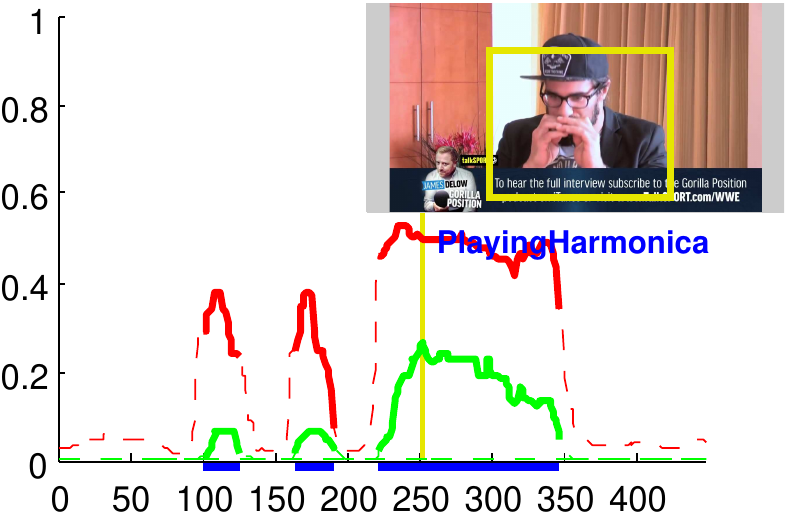}}}
         \\ \\        
         
         \includegraphics[width=55mm,height=\figy]{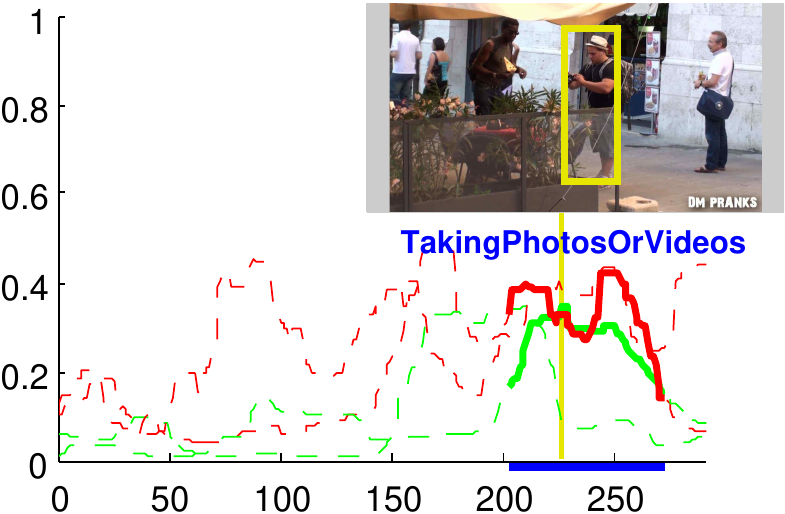}
         & \includegraphics[width=55mm,height=\figy]{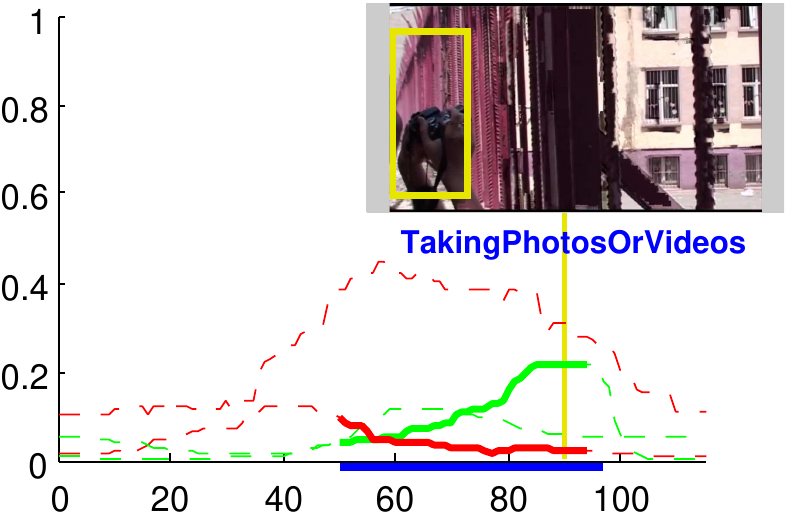}
         &  \color{red}{\fbox{\includegraphics[width=55mm,height=\figy]{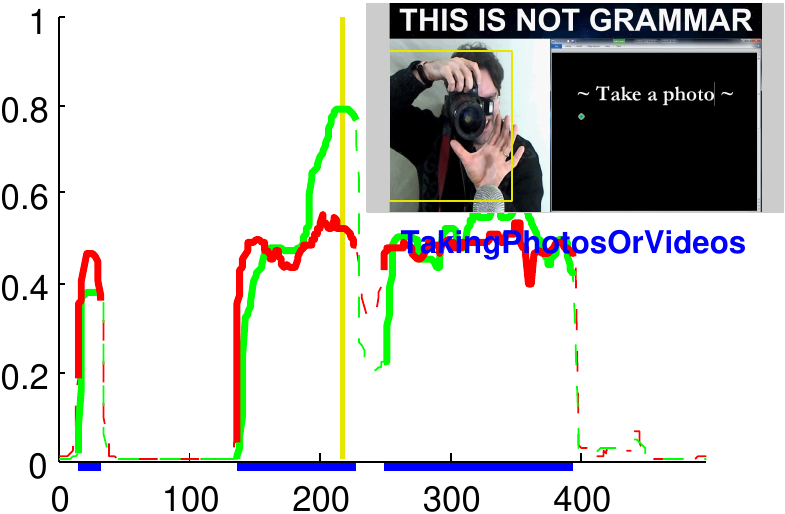}}}
    \end{tabular}
    \caption{Qualitative results for temporal localization on DALY. Each curve
      represents a human track, green curves correspond to our RecLNet
      scores and red ones to scores from CNN features~\emph{CNN RGB+OF}. A
      bold curve (resp. dashed curve) shows that the track overlaps
      with more (resp. less) than $0.5$ spatial IoU with a
      ground-truth action. The horizontal blue bar represents the
      ground-truth time segment. The video frame corresponds to the
      maximum RecLNet response (localized at the vertical yellow line)
      with its associated human detection (yellow box). The last column (red boxes) corresponds to failure cases or examples where our RecLNet model does not improve temporal localization. Each row corresponds to one action class, namely \emph{Folding Textile}, \emph{Ironing}, \emph{Phoning}, \emph{Playing Harmonica } and \emph{Taking Photos Or Videos}. The x-axis represents frame numbers.
    \label{scoreblobs2}}
\end{figure*}

\bibliographystyle{abbrvnat}
\bibliography{egbib}

\end{document}